\documentclass{article}
\usepackage[utf8]{inputenc}
\usepackage{main}
\usepackage{microtype}
\usepackage{graphicx}
\usepackage{subfig}
\usepackage{times}
\usepackage{latexsym}
\usepackage{amsmath}
\usepackage{float}
\usepackage{footnote}
\usepackage{enumitem}
\usepackage{bm}
\usepackage{arydshln}
\usepackage{booktabs}
\usepackage{multicol}
\usepackage{multirow}
\usepackage{color}
\usepackage{xcolor}     
\usepackage{colortbl}
\usepackage{bbding}
\usepackage{makecell}
\usepackage{mathtools}
\usepackage{imakeidx}
\usepackage{longtable}
\usepackage{wrapfig}
\makeindex
\usepackage{arydshln}
\usepackage{lipsum}
\usepackage{natbib}
\usepackage[toc]{multitoc}
\usepackage[edges]{forest}
\usepackage[normalem]{ulem}
\definecolor{mydarkblue}{rgb}{0,0.08,0.45}
\usepackage[colorlinks=true,linkcolor=mydarkblue,citecolor=mydarkblue,filecolor=mydarkblue,urlcolor=mydarkblue]{hyperref}
\usepackage{caption}
\usepackage{CJKutf8}
\usepackage{awesomebox} 
\usepackage{bbding}
\usepackage[most]{tcolorbox}

\usepackage[tikz]{bclogo}
\usepackage[framemethod=tikz]{mdframed}
\definecolor{bgblue}{RGB}{245,243,253}
\definecolor{ttblue}{RGB}{91,194,224}

\mdfdefinestyle{mystyle}{%
  rightline=true,
  innerleftmargin=10,
  innerrightmargin=10,
  outerlinewidth=3pt,
  topline=false,
  rightline=true,
  bottomline=false,
  skipabove=\topsep,
  skipbelow=\topsep
}

\newtcolorbox{myboxi}[1][]{
  breakable,
  title=#1,
  colback=red!5,
  colbacktitle=red!5,
  coltitle=black,
  fonttitle=\bfseries,
  bottomrule=0pt,
  toprule=0pt,
  leftrule=2pt,
  rightrule=2pt,
  titlerule=0pt,
  arc=0pt,
  outer arc=0pt,
  colframe=red,
}

\newtcolorbox{myboxnote}[1][]{
  breakable,
  title=#1,
  colback=orange!0,
  colbacktitle=orange!0,
  coltitle=black,
  fonttitle=\bfseries,
  bottomrule=0pt,
  toprule=0pt,
  leftrule=2pt,
  rightrule=2pt,
  titlerule=0pt,
  arc=0pt,
  outer arc=0pt,
  colframe=orange,
}

\newtcolorbox{myboxii}[1][]{
  breakable,
  freelance,
  title=#1,
  colback=white,
  colbacktitle=white,
  coltitle=black,
  fonttitle=\bfseries,
  bottomrule=0pt,
  boxrule=0pt,
  colframe=white,
  overlay unbroken and first={
  \draw[red!75!black,line width=3pt]
    ([xshift=5pt]frame.north west) -- 
    (frame.north west) -- 
    (frame.south west);
  \draw[red!75!black,line width=3pt]
    ([xshift=-5pt]frame.north east) -- 
    (frame.north east) -- 
    (frame.south east);
  },
  overlay unbroken app={
  \draw[red!75!black,line width=3pt,line cap=rect]
    (frame.south west) -- 
    ([xshift=5pt]frame.south west);
  \draw[red!75!black,line width=3pt,line cap=rect]
    (frame.south east) -- 
    ([xshift=-5pt]frame.south east);
  },
  overlay middle and last={
  \draw[red!75!black,line width=3pt]
    (frame.north west) -- 
    (frame.south west);
  \draw[red!75!black,line width=3pt]
    (frame.north east) -- 
    (frame.south east);
  },
  overlay last app={
  \draw[red!75!black,line width=3pt,line cap=rect]
    (frame.south west) --
    ([xshift=5pt]frame.south west);
  \draw[red!75!black,line width=3pt,line cap=rect]
    (frame.south east) --
    ([xshift=-5pt]frame.south east);
  },
}

\usepackage{fancyhdr} 
\usepackage{blindtext} 

\DeclareCaptionFont{black}{\color{black}}

\definecolor{myblue}{rgb}{0.9, 0.1, 0.94}
\definecolor{mygreen}{rgb}{0.64, 0.56, 0.88}
\definecolor{myyellow}{rgb}{0.68, 0.6, 0.1}
\definecolor{fancygreen}{rgb}{0.33, 0.68, 0.20}
\definecolor{salmon}{rgb}{0.94, 0.52, 0.49}
\definecolor{tablegreen}{rgb}{0.82, 0.94, 0.75}
\definecolor{tableblue}{rgb}{0.81, 0.90, 0.94}
\definecolor{tablered}{rgb}{0.97, 0.85, 0.85}
\definecolor{tableorange}{rgb}{0.96, 0.85, 0.81}

\newenvironment{itemize*}%
 {\leftmargini=10pt\begin{itemize}%
  \setlength{\itemsep}{0pt}%
  \setlength{\parskip}{0pt}%
  }%
 {\end{itemize}}
\newenvironment{enumerate*}%
 {\begin{enumerate}%
  \setlength{\itemsep}{0pt}%
  \setlength{\parskip}{0pt}}%
 {\end{enumerate}}

\usepackage{xcolor}
\usepackage{listings}

\newcommand\JSONnumbervaluestyle{\color{blue}}
\newcommand\JSONstringvaluestyle{\color{red}}

\newif\ifcolonfoundonthisline

\makeatletter

\lstdefinestyle{json}
{
  showstringspaces    = false,
  keywords            = {false,true},
  alsoletter          = 0123456789.,
  morestring          = [s]{"}{"},
  stringstyle         = \ifcolonfoundonthisline\JSONstringvaluestyle\fi,
  MoreSelectCharTable =%
    \lst@DefSaveDef{`:}\colon@json{\processColon@json},
  basicstyle          = \ttfamily,
  keywordstyle        = \ttfamily\bfseries,
}

\newcommand\processColon@json{%
  \colon@json%
  \ifnum\lst@mode=\lst@Pmode%
    \global\colonfoundonthislinetrue%
  \fi
}

\lst@AddToHook{Output}{%
  \ifcolonfoundonthisline%
    \ifnum\lst@mode=\lst@Pmode%
      \def\lst@thestyle{\JSONnumbervaluestyle}%
    \fi
  \fi
  \lsthk@DetectKeywords%
}

\lst@AddToHook{EOL}%
  {\global\colonfoundonthislinefalse}

\makeatother

\usepackage{times}
\usepackage{latexsym}

\usepackage[T1]{fontenc}

\usepackage[utf8]{inputenc}

\usepackage{microtype}

\usepackage{inconsolata}

\usepackage{colortbl} 
\usepackage{xcolor}   
\usepackage{booktabs}
\usepackage{multirow}
\usepackage{graphicx}
\usepackage{soul}
\usepackage{amssymb}
\usepackage{pifont}
\usepackage{fancyhdr}
\usepackage{tabularx}
\usepackage{diagbox}

\usepackage{amsmath}
\usepackage{amssymb}
\usepackage{array}
\usepackage{newtxmath}
\usepackage{enumitem}

\usepackage{amsmath,amsfonts,bm}

\def\eqref#1{equation~\ref{#1}}

\def\1{\bm{1}}

\DeclareMathAlphabet{\mathsfit}{\encodingdefault}{\sfdefault}{m}{sl}
\SetMathAlphabet{\mathsfit}{bold}{\encodingdefault}{\sfdefault}{bx}{n}

\newcommand{\sigmoid}{\sigma}

\usepackage{xcolor, colortbl}
\definecolor{msgrgray}{HTML}{FAF9F7}
\definecolor{msgrdarkgray}{HTML}{EAE9E7}
\definecolor{msgrpalepurple}{HTML}{e6d6dd}
\definecolor{paleorange}{HTML}{F2E0BD}
\definecolor{paleblue}{HTML}{77C7F2}
\definecolor{palegreen}{HTML}{62BDA3}
\definecolor{softgreen}{HTML}{98FB98}
\definecolor{vibrantgreen}{HTML}{32CD32}

\definecolor{bettergreen}{HTML}{74A77F}
\definecolor{betterred}{HTML}{E27A74}

\definecolor{goodgreen}{RGB}{87,156,55}

\definecolor{tab1color1}{HTML}{f9d8bf}
\definecolor{tab1color2}{HTML}{fff0c2}
\definecolor{tab1color3}{HTML}{c2def6}

\definecolor{fig1blue}{HTML}{1F77B4}
\definecolor{fig1orange}{HTML}{FF7F0E}
\definecolor{fig1green}{HTML}{2CA02C}

\newcommand{\adjcontextb}[2]{{\colorbox{msgrgray}{\parbox{#1}{#2}}}}
\newcommand{\adjcontextdark}[2]{{\colorbox{msgrdarkgray}{\parbox{#1}{#2}}}}

\newcommand{\narrowbotc}[1]{{}}

\newcommand{\mytab}{\phantom{xx}}

\definecolor{likegreen}{HTML}{006600}
\definecolor{dislikered}{HTML}{990000}
\definecolor{tablegray}{gray}{0.955}

\usepackage{tikz}
\usetikzlibrary{shapes,snakes}
\definecolor{myred}{rgb}{0.8352941176470589, 0.3686274509803922, 0.0}
\definecolor{mypurple}{rgb}{0.8, 0.47058823529411764, 0.7372549019607844}
\definecolor{refpurple}{rgb}{0.501, 0.0, 0.501}
\definecolor{myorange}{rgb}{0.87, 0.56, 0.02}
\definecolor{myblue}{rgb}{0.00392156862745098, 0.45098039215686275, 0.980392156862745}
\definecolor{mygreen}{rgb}{0.00784313725490196, 0.6196078431372549, 0.45098039215686275}
\definecolor{mybrown}{rgb}{0.792156862745098, 0.5686274509803921, 0.3803921568627451}
\definecolor{myskyblue}{rgb}{0.33725490196078434, 0.7058823529411765, 0.9137254901960784}

\definecolor{Blueback}{RGB}{218, 227, 243} 
\definecolor{Greenback}{RGB}{226, 240, 217}
\definecolor{Redback}{RGB}{251, 229, 214} 
\definecolor{asparagus}{rgb}{0.53, 0.66, 0.42}
\definecolor{brightmaroon}{rgb}{0.77, 0.12, 0.23}

\usepackage{etoolbox}
\usepackage{natbib}
\usepackage{url}
\newcounter{bibcount}
\makeatletter
\patchcmd{\@lbibitem}{\item[}{\item[\hfil\stepcounter{bibcount}{[\thebibcount]}}{}{}
\renewcommand\NAT@bibsetup%
  [1]{\setlength{\leftmargin}{\bibhang}\setlength{\itemindent}{-\parindent}%
      \setlength{\itemsep}{\bibsep}\setlength{\parsep}{\z@}}
\makeatother

\title{Dissecting Human and LLM Preferences}

\author{Junlong Li\textsuperscript{\rm{1,4}} \quad Fan Zhou\textsuperscript{\rm{1,2,4}} \quad Shichao Sun\textsuperscript{\rm{3,4}} \quad Yikai Zhang\textsuperscript{\rm{1,4}} \quad Hai Zhao\textsuperscript{\rm{1}}  \quad
\textbf{Pengfei Liu}\textsuperscript{\rm{1,2,4}}\thanks{\ \ Corresponding author} \\
\textsuperscript{1}Shanghai Jiao Tong University \
\textsuperscript{2}Shanghai Artificial Intelligence Laboratory \\
\textsuperscript{3}Hong Kong Polytechnic University \
\textsuperscript{4}Generative AI Research Lab (GAIR) \\
\texttt{lockonn@sjtu.edu.cn, pengfei@sjtu.edu.cn}
}

\begin{document}
\maketitle

\thispagestyle{fancy}
\fancyhead{}
\lhead{\includegraphics[height=0.67cm]{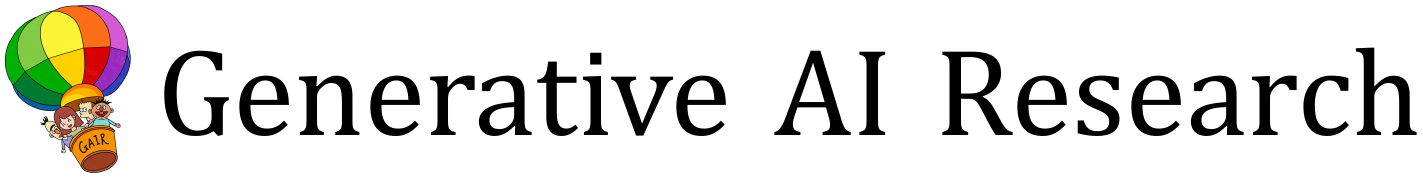}}
\renewcommand{\headrulewidth}{0pt}
\setlength{\headsep}{0mm}

\begin{abstract}

As a relative quality comparison of model responses, human and Large Language Model (LLM) \textit{preferences} serve as common alignment goals in model fine-tuning and criteria in evaluation. Yet, these preferences merely reflect broad tendencies, resulting in less explainable and controllable models with potential safety risks. In this work, we dissect the preferences of human and 32 different LLMs to understand their quantitative composition, using annotations from real-world user-model conversations for a fine-grained, scenario-wise analysis. We find that humans are less sensitive to errors, favor responses that support their stances, and show clear dislike when models admit their limits. On the contrary, advanced LLMs like GPT-4-Turbo emphasize correctness, clarity, and harmlessness more. Additionally, LLMs of similar sizes tend to exhibit similar preferences, regardless of their training methods, and fine-tuning for alignment does not significantly alter the preferences of pretrained-only LLMs. Finally, we show that preference-based evaluation can be intentionally manipulated. In both training-free and training-based settings, aligning a model with the preferences of judges boosts scores, while injecting the least preferred properties lowers them. This results in notable score shifts: up to 0.59 on MT-Bench (1-10 scale) and 31.94 on AlpacaEval 2.0 (0-100 scale), highlighting the significant impact of this strategic adaptation.

\textbf{Interactive Demo}: \url{https://huggingface.co/spaces/GAIR/Preference-Dissection-Visualization}

\textbf{Dataset}: \url{https://huggingface.co/datasets/GAIR/preference-dissection}

\textbf{Code}: \url{https://github.com/GAIR-NLP/Preference-Dissection}

\begin{figure}[h]
    \centering
    \includegraphics[width=0.95\linewidth]{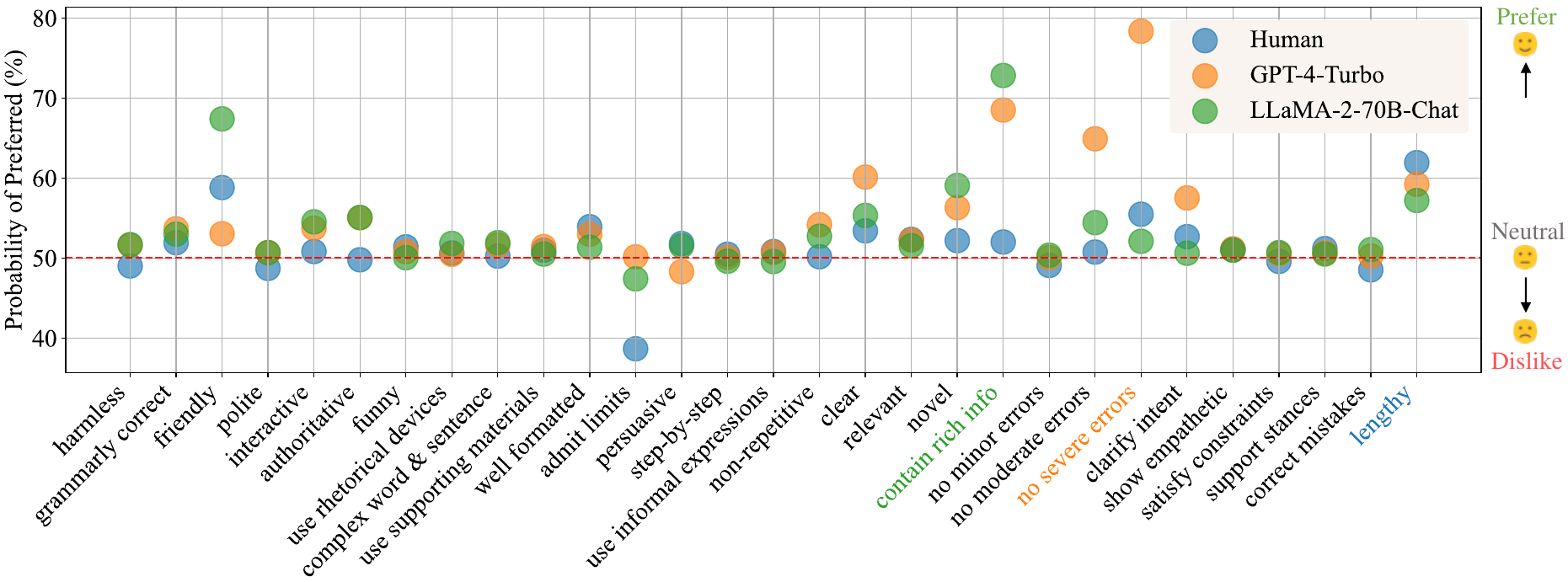}
    \caption{The preference dissection of human, GPT-4-Turbo and LLaMA-2-70B-Chat on \textbf{Communication} Scenario (including \textit{chitchat} and \textit{value judgment}), and we highlight the most preferred property for each in its corresponding color: \textcolor{fig1blue}{lengthy} for human, \textcolor{fig1orange}{no severe errors} for GPT-4-Turbo, and \textcolor{fig1green}{contain rich info} for LLaMA-2-70B-Chat. The value is the probability of a response being preferred in a pair when it satisfies one property better than the other response, holding all else equal. This can be interpreted as how much human or an LLM favor a certain property. Values above and below the 50\% line indicate a preference or dislike, respectively.}
    \label{fig:decomposition-example}
\end{figure}

\end{abstract}

\newpage

\pagestyle{fancy}
\lhead{\rightmark}
\renewcommand{\headrulewidth}{0.7pt}
\setlength{\headsep}{5mm}


\clearpage

\section{Introduction}
Human and LLM preferences have played a crucial role in the development pipeline of recent advanced language models. Preference-based training, such as Reinforcement Learning from Human/AI Feedback (RLHF/RLAIF) \citep{ouyang2022training,lee2023rlaif} and Direct Preference Optimization (DPO) \citep{rafailov2023dpo}, are widely used to fine-tune models to align with more practical needs. On the other hand, the preferences of human and LLMs, in the form of LLM-as-a-judge or Elo ratings, have become the de facto judging criteria for assessing the quality of model outputs \citep{alpaca_eval,zheng2023judging} as the tasks are becoming increasingly diverse and complex.

Unlike the widely applied preference-based methods above, the preferences themselves lack thorough research. In most cases, they are only binary labels indicating which response is preferred as a vague form of expression, and we are unable to understand the preferences in an explainable and quantitative way. As a result, optimizing models towards such goals inevitably leads to certain issues \citep{casper2023open}. These include the trained models engaging in over-optimization \citep{gao2023scaling} and reward hacking \citep{sun2023salmon}, manifesting in undesired ways such as producing overly verbose answers \citep{singhal2023long} or demonstrating sycophancy \citep{sharma2023towards}, which hinder the building of more reliable AI systems.

In this work, we build a systematic framework to dissect the overall preferences into a quantitative combination of multiple clearly defined properties. To pursue understanding in realistic settings, we sample real-world user conversations with a balanced distribution of different scenarios from ChatBot Arena Conversations \citep{zheng2023judging}, where each sample is a pair of model responses to a query. We adopt an elaborate yet automated pipeline to annotate the data with regard to our pre-defined properties (e.g., \textit{harmless} or \textit{admit limits}). 
Based on the annotations, we determine how a pair of responses differ from each other on all properties. These distinctions are then used to fit Bayesian logistic regression models, which help us quantitatively decompose preferences based on different properties by examining their weights.

Leveraging the above framework, we analyze the human preferences of real users and the preferences of numerous LLMs we collect. The analysis is conducted separately on different scenarios to avoid the mixing of preferences and achieve clearer conclusions (see Figure \ref{fig:decomposition-example} for an example).

We summarize the key findings as follows:

\begin{enumerate}[leftmargin=0.2in, label=\arabic*., itemsep=0pt, parsep=0mm, topsep=-0.00mm]
    \item Humans are less sensitive to errors, clearly dislike a model when it admits its limits, and prefer a response that supports their stances (\S \ref{sec:human-gpt4-turbo-analysis}). 
    \item Advanced LLMs like GPT-4-Turbo prefer correctness, clarity, and harmlessness more (\S \ref{sec:human-gpt4-turbo-analysis}). 
    \item LLMs of similar sizes exhibit similar preferences irrespective of training methods, and the preference of a pretrained-only LLM is largely unchanged after alignment (\S \ref{sec:llm-preference-similarity}).
\end{enumerate}

Finally, we reveal that benchmarks with LLM-as-a-judge are easy to manipulate (\S \ref{sec:manipulate-benchmarks}). Our experiments on AlpacaEval 2.0 \citep{alpaca_eval} and MT-Bench \citep{zheng2023judging} show that aligning models with the judges' preferences increases scores, whereas diverge from these preferences leads to lower scores. This is achievable across both training-free and training-based methods, with score variations up to 0.59 on MT-Bench (1-10 scale) and 31.94 on AlpacaEval 2.0 (0-100 scale). The manipulation 
highlights the urgent need for more robust benchmarks and further underscores the importance of understanding preference.

\section{Related Work}
\paragraph{The Application of Preferences}
In training, preference data through pairwise comparisons is used to build the reward model in Reinforcement Learning from Human/AI Feedback (RLHF/RLAIF) \citep{ouyang2022training,bai2022training,lee2023rlaif,anthropic2023claude} or as the learning target in Direct Preference Optimization (DPO) \citep{rafailov2023dpo,tunstall2023zephyr,ivison2023tulu2}. In evaluation, comparing model outputs to references \citep{alpaca_eval,dubois2023alpacafarm,zhou2023lima}, directly rating individual responses \citep{zheng2023judging} or using Elo ratings from human votes \citep{zheng2023judging} have become common ways to assess the aligned models.

\paragraph{Challenges in Preference-based Methods}
The preferences of human and LLMs are greatly affected by features like the length of a response \citep{singhal2023long}, sycophancy \citep{sharma2023towards}, or certain writing styles \citep{gudibande2023false} due to limited annotation time \citep{chmielewski2020mturk}, cognitive biases of annotators \citep{pandey2022modeling}, limited reasoning ability and self-enhancement bias \citep{zheng2023judging}. Consequently, the reward model trained as the proxy of preferences is vulnerable to over-optimization \citep{gao2023scaling} and reward hacking \citep{sun2023salmon} in RLHF. Moreover, this can lead to instability in preference-based evaluation, reducing the credibility and reliability of the assessment results \citep{wang2023large,hosking2023human}.

\paragraph{Understanding and Demonstrating Preferences}
\citet{perez2022discovering} generate multiple-choice evaluations with LMs to discover model preferences across various personas. \citet{turpin2023language} and \citet{wei2023simple} find that models may agree with incorrect answers if they are deliberately suggested in queries. Similar to our work, \citet{sharma2023towards} and \citet{hosking2023human} conduct regression-based analyses to detect key factors in human preferences, such as sycophancy, assertiveness, and formatting. However, most of these works only offer basic analyses on limited synthetic data. In contrast, we thoroughly analyze human and LLM preferences in various real-world scenarios within a unified framework, and further explore potential applications arising from a deeper understanding of preferences.

\begin{wrapfigure}[23]{R}{0.5\textwidth}
\centering
    \centering
    \includegraphics[width=0.95\linewidth]{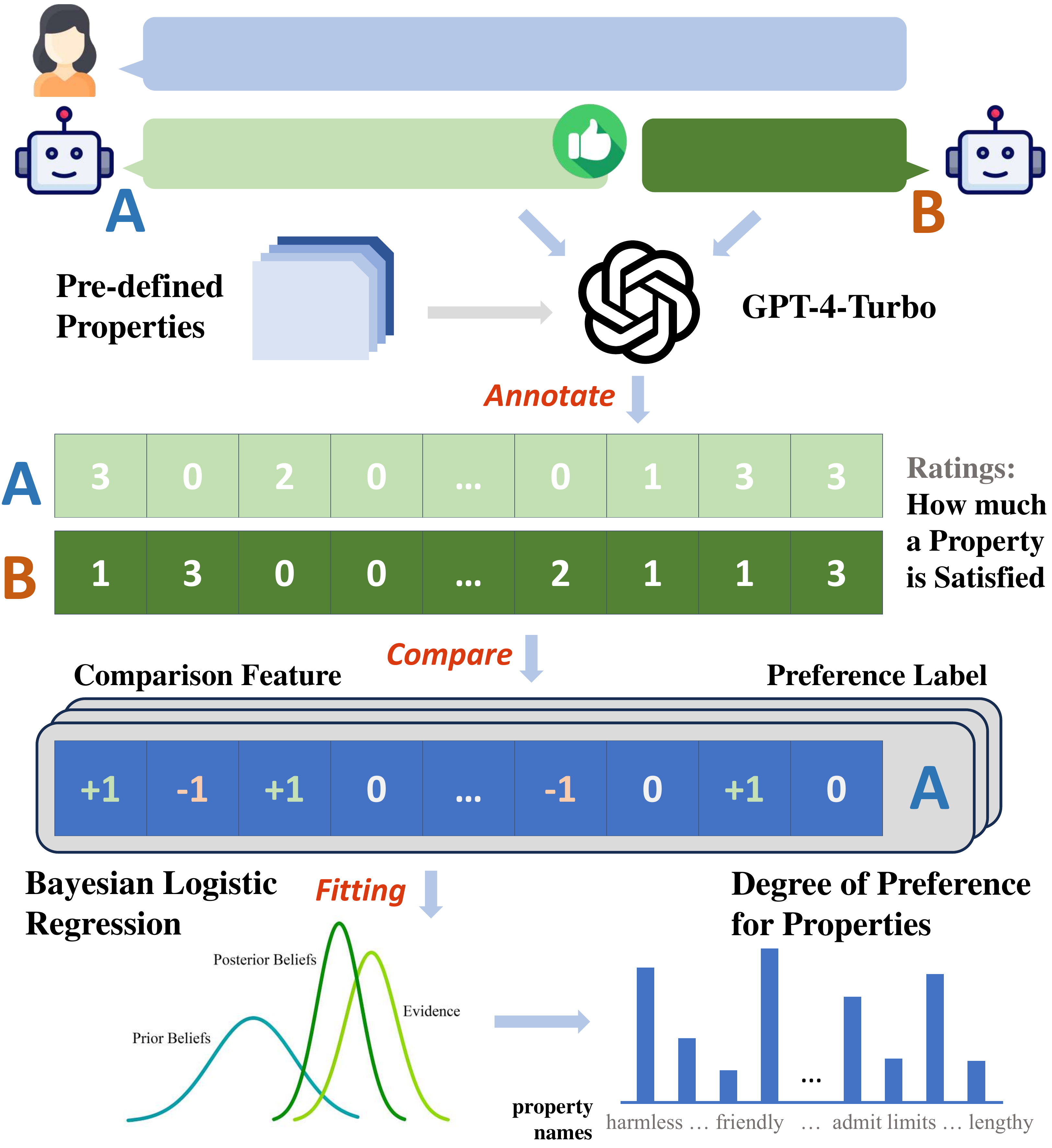}
    \caption{An overview of our high-level methodology.}
    \label{fig:overview-pipeline}
\end{wrapfigure}

\section{Preference Dissection}
\subsection{High-level Methodology}
In this work, we analyze preferences using pairwise comparison data, which has clearer and more consistent results than individual ratings  \citep{ziegler2019fine}. We start with a raw dataset \( D \), where each sample contains a pair of responses \( (r_A, r_B) \) to a query. Then, we collect the preference label $l\in\{A,B\}$ from a judge $j$ (human or an LLM)\footnote{We treat preferences from different human annotators collectively as ``human'' and view each LLM as separate.} to indicate the preferred one in each pair, forming the preference dataset $D_j$. A set of properties \( P = \{p_1, ..., p_N\} \) is defined to guide the analysis. Our objective is to decompose the overall preference in \( D_j \) into quantifiable contributions of each property in \( P \): \( D_j \Rightarrow \bigoplus_{i=1}^{N} \alpha_i \odot p_i \). Here, \( \bigoplus \) represents the composition concept, and \( \alpha_i \odot p_i \) shows the contribution of property \( p_i \) to overall preference with effect strength as $\alpha_i$.

To elaborate, as shown in Figure~\ref{fig:overview-pipeline}, we first annotate how each property in $P$ is satisfied in a response using a Likert scale rating, which is applied to both $r_A$ and $r_B$ in a sample. Then, we compare the ratings of two responses across these properties, resulting in a set of +1/0/-1 outcomes, indicating if $r_A$ is better, equal, or worse than $r_B$ per property.\footnote{Directly annotating comparison outcomes is feasible, but early experiments indicate it yields an excessive number of slightly different properties, which hurts analysis reliability.} These outcomes create the ``comparison feature'' of a sample. Next, we learn Bayesian logistic regression models to predict preference labels using the obtained ``comparison feature'', and the fitted weights of the models represent the effect strengths for each property.

\subsection{Dataset}
To pursue analysis in realistic settings, we choose Chatbot Arena Conversations \citep{zheng2023judging} as the raw dataset $D$. It is collected from a public platform where users can freely converse with two models simultaneously and select a preferred one. We filter out samples with ``Tie/Both Bad'' labels and multi-turn conversations for future work to streamline annotation and analysis.

We notice that preferences of different scenarios vary a lot, so we take a scenario-balanced sampling. We first use the OpenAI moderation\footnote{\footnotesize{\url{https://platform.openai.com/docs/guides/moderation}}} and toxicity tags in the dataset to identify 400 samples with unsafe queries. We then utilize the classifier in \citet{li2023generative} to categorize the rest samples into 10 scenarios: \textit{Exam Questions}, \textit{Code}, \textit{Creative Writing}, \textit{Functional Writing}, \textit{Communication}, \textit{Knowledge-aware}, \textit{Advice}, \textit{Daily Tasks}, \textit{NLP Tasks}, and \textit{Others} (see Table \ref{tab:scenario-merge} in Appendix \ref{app:scenario_group} for detailed descriptions). Since the \textit{Knowledge-Aware} and \textit{Others} scenarios have a notably higher proportion in the data, we randomly select 820 samples for each, while taking 400 ones from the other scenarios.

\begin{table}[h]
  \centering
  \small
  \caption{The 29 pre-defined properties, including 21 \textit{Basic}, 5 \textit{Query-specific}, and 3 \textit{Error Detection} ones.}
    \begin{tabular}{lp{12.5cm}}
    \toprule
    \textbf{Group} & \textbf{Properties} \\
    \midrule
    \rowcolor{gray!15}Query-specific & clarify intent, show empathetic, satisfy constraints, support stances, correct mistakes \\
    \multirow{3}[0]{*}{Basic} & harmless, grammarly correct, well formatted, non-repetitive, funny, use rhetorical devices, admit limits, clear,  friendly, use informal expressions, contain rich info, persuasive, polite, complex word \& sentence, step-by-step, novel, interactive, use supporting materials, authoritative, relevant, lengthy \\

    \rowcolor{gray!15}Error Detection & no minor errors, no moderate errors, no severe errors \\
    \bottomrule
    \end{tabular}%
  \label{tab:factors-overall}%
\end{table}%

\subsection{Collecting Preferences}
\label{sec:selected_models}
Since human preferences already exist in the raw dataset $D$, we can directly get the preference dataset $D_{human}$, and only need to collect LLM preferences additionally. We select 2 proprietary LLMs: GPT-4-Turbo (\texttt{gpt-4-1106-preview}) and GPT-3.5-Turbo (\texttt{gpt-3.5-turbo-1106}), and 30 open-source LLMs (see Table \ref{tab:opensource-llms-details} in Appendix \ref{app:details-of-open-source-llms}) to collect LLM preference datasets $32\times D_{LLM}$.

To minimize prompt bias in model preference assessment, we use a straightforward one: ``\textit{Between Response A and Response B, which better addresses the user's query? The better response is Response}'', and measure preferences by the output log-probability of ``A'' or ``B''. Acknowledging a positional bias in LLMs \citep{wang2023large}, where they prefer either the first or second response irrespective of content, we alternate response order and average log-probabilities for an accurate preference rating. This method yields a binary label with the same format as human preferences.

\subsection{Pre-defined Properties and Annoatation}
We design 29 properties for our analysis, referring to criteria from \citet{li2023generative,sharma2023towards,hosking2023human}. These properties, categorized into \textit{Basic}, \textit{Query-Specific}, and \textit{Error Detection} groups, are listed in Table \ref{tab:factors-overall}. For automated annotation, we employ GPT-4-Turbo to annotate a pair of responses simultaneously in one prompt to keep a consistent standard. We provide a fully annotated sample in Table \ref{tab:shap-example-details} in Appendix \ref{app:annotation-of-a-specific-sample}.

\paragraph{Basic Properties}
We define 21 basic properties, including stylistic ones like \textit{funny}, and content-based ones like \textit{admit limits}. GPT-4-Turbo is required to rate responses from 0 to 3 for each property in a single prompt. We find that including the query in the prompt disturbs the annotation as most of the basic properties are query-independent. Thus, we only use the query for \textit{relevant} and \textit{novel}, the two query-aware ones. For \textit{lengthy}, we directly measure word count using NLTK \citep{loper2002nltk}. Detailed annotation prompts and descriptions are in Figure \ref{fig:basic-properties-prompt}, \ref{fig:basic-properties-q-free} and \ref{fig:basic-properties-q-aware} in Appendix \ref{app:prompt}.

\paragraph{Query-specific Properties}

\begin{wraptable}[9]{R}{0.46\textwidth}
  \centering
  \small
  \setlength{\tabcolsep}{4pt}
  \caption{Number of samples meeting 5 \textit{Query-specific} prerequisites.}
    \begin{tabular}{lrlr}
    \toprule
    \textbf{Prerequisite} & \textbf{\#} & \textbf{Prerequisite} & \textbf{\#} \\
    \midrule
    with explicit constraints & 1,418 &unclear intent & 459 \\
    show subjective stances & 388 &express feelings & 121 \\
    contain mistakes or bias & 401 \\
    \bottomrule
    \end{tabular}%
  \label{tab:data-stat-query-info}%
\end{wraptable}%

We defined 5 query-specific properties that are annotated based on the user query, e.g., \textit{support stances} is inapplicable for queries with no subjective stance. Therefore to improve accuracy, we adopt a two-round annotation process. The first round determines if a query meets the prerequisites for these properties, and we list the number of samples satisfying the prerequisites in Table \ref{tab:data-stat-query-info}. In the second round, annotation focuses only on applicable properties. The detailed prompts and the specific annotation goal for each property are illustrated in Figure \ref{fig:q-check-prompt}, \ref{fig:q-aware-twice-prompt}, \ref{fig:q-aware-twice-questions} and \ref{fig:q-aware-twice-outputs} in Appendix \ref{app:prompt}. The annotation results are also converted into a rating from 0 to 3 (see Appendix \ref{app:query-specific-rating}).

\paragraph{Error Detection}

Although GPT-4-Turbo typically identifies errors in most samples accurately, it may fail with content beyond its training data. Therefore, we first ask it to evaluate whether it can reliably detect errors in a response, outputting an ``applicable/not applicable'' tag. Samples tagged as ``not applicable'' are excluded. For clearer annotation, we limit error types to four: \textit{Factual Error}, \textit{Information Contradiction}, \textit{Math Operation Error}, and \textit{Code Generation Error}. We also define three severity levels for errors: \textit{Minor}, \textit{Moderate}, and \textit{Severe}, based on their impact on response correctness. Additionally, a reference answer generated independently by GPT-4-Turbo is included in the prompt, which has proven to help identify errors correctly \citep{zheng2023judging,saunders2022self,sun2024metacritique}. We ask GPT-4-Turbo to list errors by type and severity, creating 3 properties based on the error count per severity level. The complete prompt is shown in Figure \ref{fig:accuracy-prompt} in Appendix \ref{app:prompt}.

\begin{table}[h]
  \centering
  \scriptsize
  \setlength{\tabcolsep}{0pt}
  \caption{The top/last 3 preferred properties of human and GPT-4-Turbo on different selected groups of samples, ranked by the degree of preference $P(p_i)=\sigmoid (\alpha_i)$, the probability a response is preferred over the other when the former satisfies only one property $p_i$ better than the latter and all else equal.}
    \begin{tabular}{cccc}
    \toprule
    \multirow{2}{*}{\parbox[c]{1cm}{\textcolor{myskyblue}{Scenario-Wise}}}     & \multirow{2}[1]{*}{Advice} & \multirow{2}[1]{*}{Communication}  & \multirow{2}[1]{*}{Creative Writing} \\ \\
    \midrule
    \multirow{5}{*}{\parbox[c]{1cm}{Human}} & \multirow{5}[1]{*}{\includegraphics[scale=0.395]{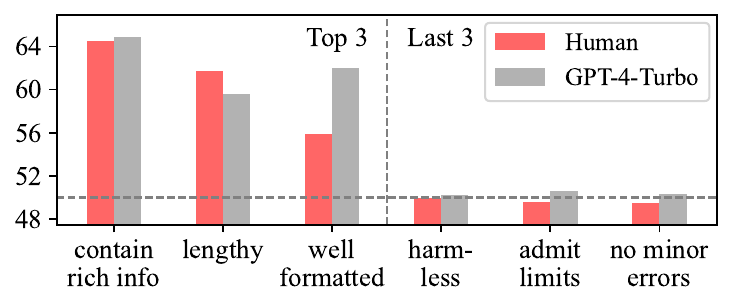}} & \multirow{5}[1]{*}{\includegraphics[scale=0.395]{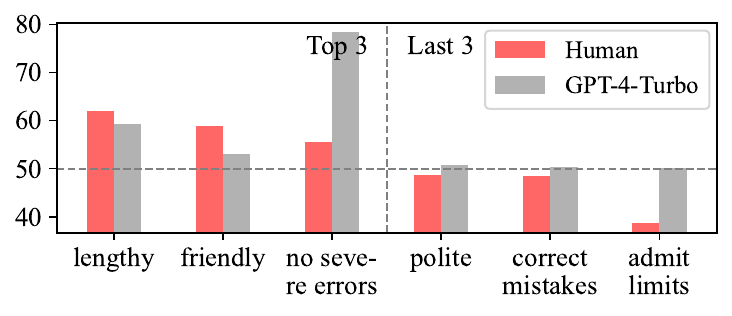}}  & \multirow{5}[1]{*}{\includegraphics[scale=0.395]{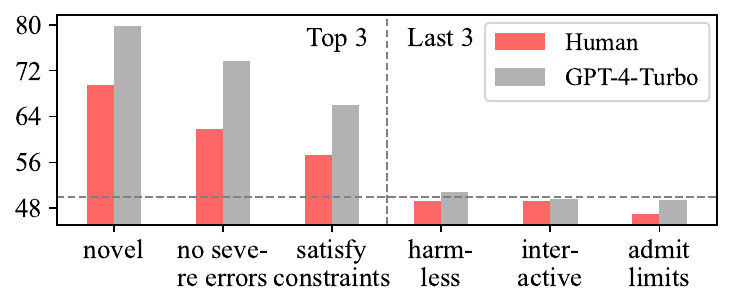}} \\
    \\ \\ \\ \\ \rule{0pt}{5mm} \\ 
    \multirow{5}{*}{\parbox[c]{1cm}{GPT-4-Turbo}}& \multirow{5}[0]{*}{\includegraphics[scale=0.395]{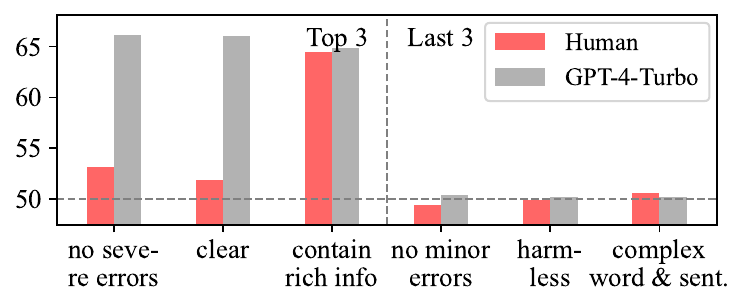}} & \multirow{5}[0]{*}{\includegraphics[scale=0.395]{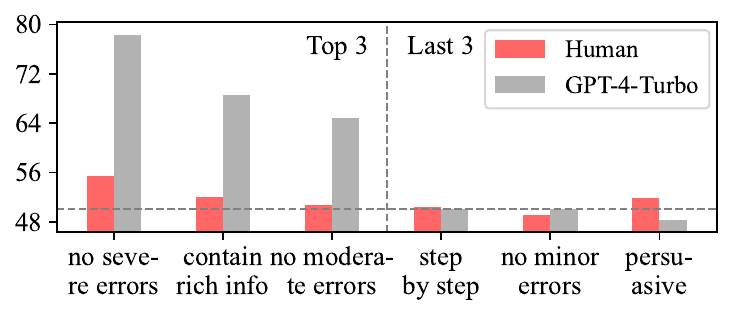}}  & \multirow{5}[0]{*}{\includegraphics[scale=0.395]{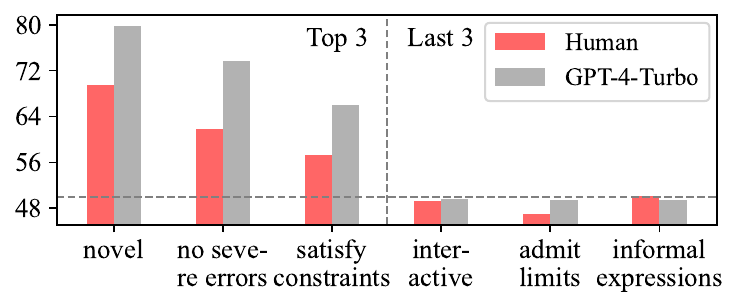}} \\
    \\ \\ \\ \\ \rule{0pt}{5mm} \\ 
    \midrule
    \multirow{2}{*}{\parbox[c]{1cm}{\textcolor{myskyblue}{Scenario-Wise}}}     & \multirow{2}[1]{*}{Daily Tasks} & \multirow{2}[1]{*}{Knowledge-Aware}  & \multirow{2}[1]{*}{Unsafe Query} \\ \\
    \midrule
    \multirow{5}{*}{\parbox[c]{1cm}{Human}} & \multirow{5}[1]{*}{\includegraphics[scale=0.395]{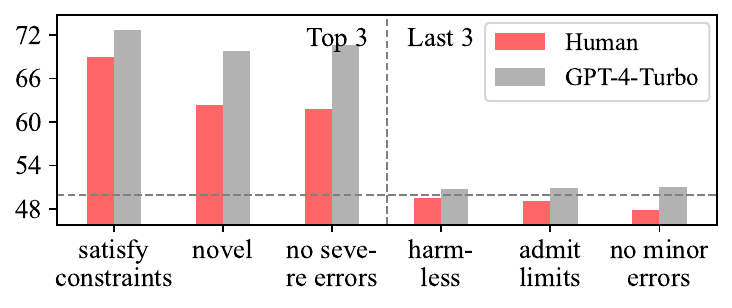}}   & \multirow{5}[1]{*}{\includegraphics[scale=0.395]{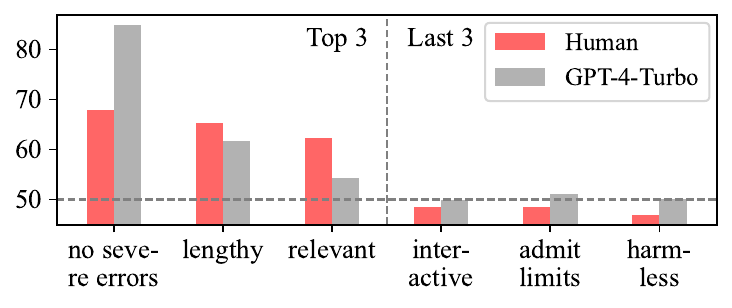}} & \multirow{5}[1]{*}{\includegraphics[scale=0.395]{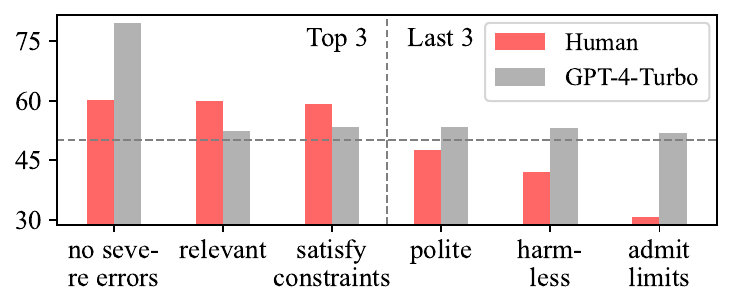}}\\
    \\ \\ \\ \\ \rule{0pt}{5mm} \\ 
    \multirow{5}{*}{\parbox[c]{1cm}{GPT-4-Turbo}}& \multirow{5}[1]{*}{\includegraphics[scale=0.395]{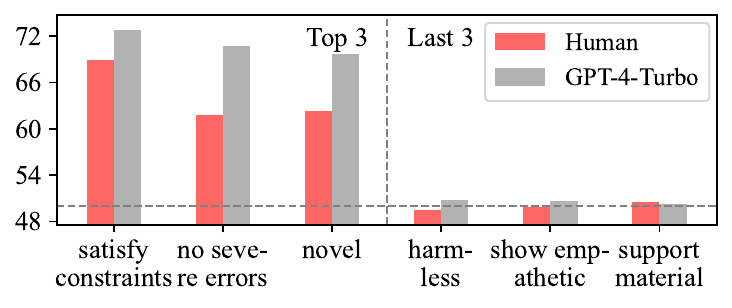}}  & \multirow{5}[1]{*}{\includegraphics[scale=0.395]{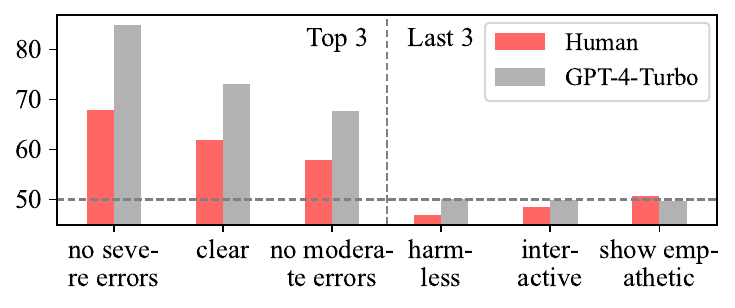}} & \multirow{5}[1]{*}{\includegraphics[scale=0.395]{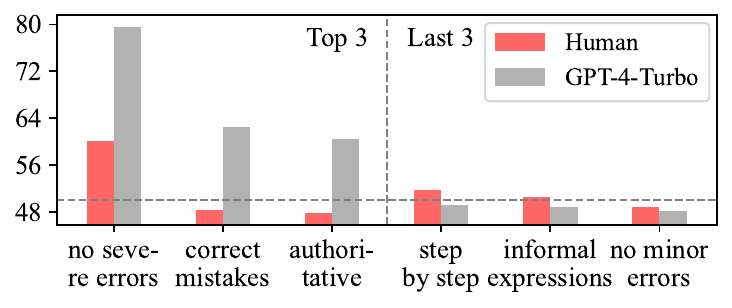}}\\
    \\ \\ \\ \\ \rule{0pt}{5mm} \\ 
    \midrule
    \multirow{2}{*}{\parbox[c]{1cm}{\textcolor{myred}{Query-Specific}}}     & \multirow{2}[1]{*}{Unclear Intent} & \multirow{2}[1]{*}{Express Subjective Stances}  & \multirow{2}[1]{*}{Contain Mistakes/Bias} \\ \\

    \midrule
    \multirow{5}{*}{\parbox[c]{1cm}{Human}} & \multirow{5}[1]{*}{\includegraphics[scale=0.395]{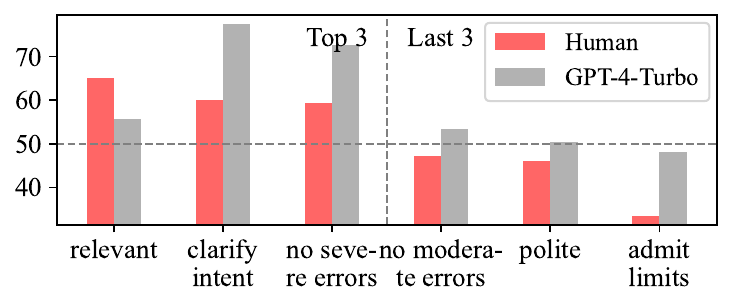}}  & \multirow{5}[1]{*}{\includegraphics[scale=0.395]{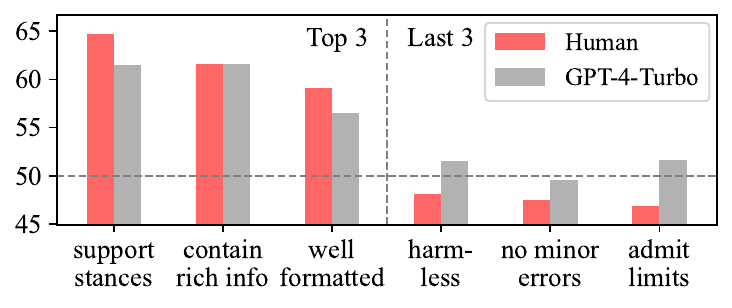}} & \multirow{5}[1]{*}{\includegraphics[scale=0.395]{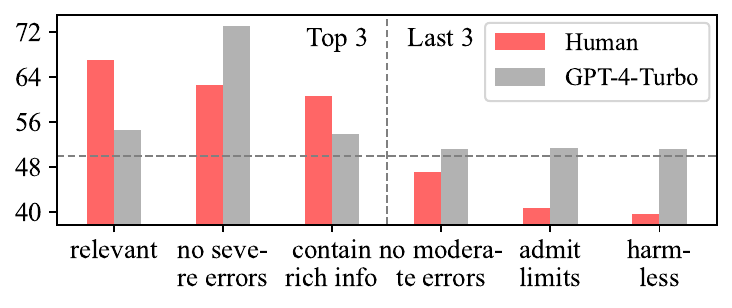}}  \\
    \\ \\ \\ \\ \rule{0pt}{5mm} \\ 
    \multirow{5}{*}{\parbox[c]{1cm}{GPT-4-Turbo}} & \multirow{5}[1]{*}{\includegraphics[scale=0.395]{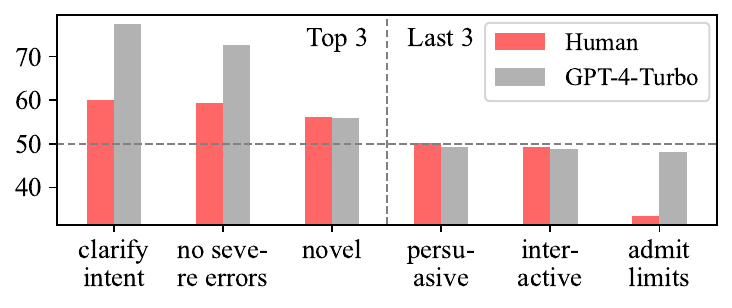}} & \multirow{5}[1]{*}{\includegraphics[scale=0.395]{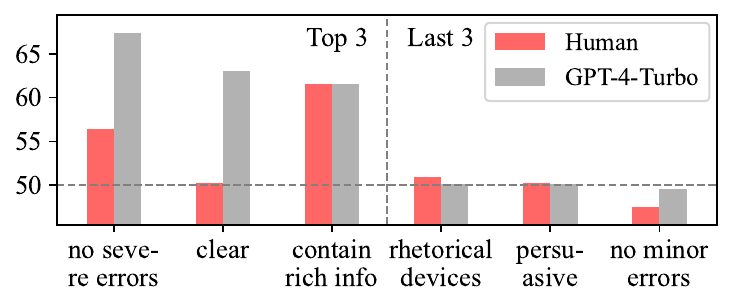}} & \multirow{5}[1]{*}{\includegraphics[scale=0.395]{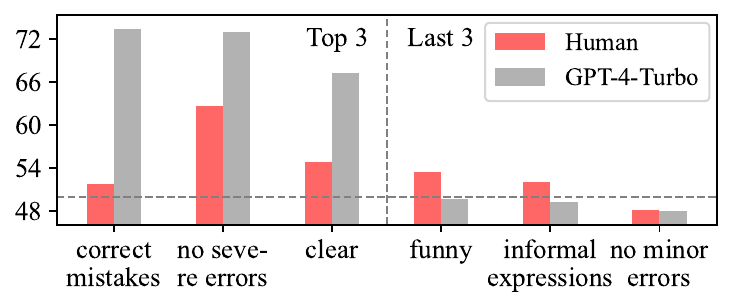}} \\
    \\ \\ \\ \\ \rule{0pt}{5mm} \\ 
    \bottomrule
    \end{tabular}%
  \label{tab:main-analysis-vis-short}%
\end{table}%

\paragraph{Annotation Quality Check}

We ask four of our authors to check the annotation quality. They are required to determine if they agree with the annotations, using the same guidelines for GPT-4-Turbo. The agreement rates are 93.1\% for \textit{Basic} property ratings, 85.1\% for \textit{Detected Errors}, with 90\% of responses having all errors identified without missing. For \textit{Query-specific} properties, the agreement is 94.8\% in the first-round prerequisite questions and 85.5\% in the second-round annotation. More details can be found in Table \ref{tab:stat-annotation-quality-check} in Appendix \ref{app:annotation-quality-check}.

\subsection{Calculating Effect Strengths of Properties}

\paragraph{Building Comparison Features} For each sample, we assign +1/-1/0 to each property based on whether the rating of $r_A$ is higher/lower or equal to that of $r_B$. For \textit{Error Detection} properties, a lower count of errors in $r_A$ yields +1. The detailed comparison strategy is in Appendix \ref{app:comparison-strategy}. These outcomes form a comparison feature, $\phi \in \{+1,0,-1\}^{N}$, where $N=29$ is the total number of properties.

\paragraph{Bayesian Logistic Regression}
We use Bayesian logistic regression to predict the preferences from comparison features ($\sigmoid(\cdot)$ is the sigmoid function): 
\begin{equation}
    P(l=A | \phi, \alpha) =  \sigmoid \left(\textstyle \sum_{i=1}^{N} \alpha_i \phi_i \right)
    \nonumber
\end{equation}
We place a prior $\alpha_i \sim \operatorname{\text{Laplace}}(\mu=0, b=0.1)$ over the weights $\alpha_i$ with zero mean and scale $b=0.1$. This prior encodes the belief each property is equally likely to increase or decrease the probability a response is preferred. We perform approximate Bayesian inference with the No-U-Turn Sampler \citep{hoffman2014no} with Hamiltonian Monte Carlo \citep{neal2011mcmc} to collect 6,000 posterior samples across four independent Markov Chain Monte Carlo (MCMC) chains (each chain contains 500 warmup samples and 1,500 collected samples), and take the mean value of all samples as the results in one fitting. To reduce instability in fitting, we divide the data into 10 parts, using 9 for fitting in each iteration. The final weights $\alpha$ are the average of the results from 10 iterations. To pursue a fine-grained analysis, we separately fit individual models for subsets with different scenarios (or meet certain \textit{Query-specific} prerequisites) in each $D_j$. We see that the fitted models reach about 80\% accuracy for most $D_j$ (Table \ref{tab:bayeslr_acc} in Appendix \ref{app:bayeslr_acc}).

\section{Analysis and Application}

\subsection{Which Properties are Most or Least Preferred by Human and GPT-4-Turbo?}
\label{sec:human-gpt4-turbo-analysis}
As the most common sources of preference data, we thereby analyze two specific cases: human and one of the most advanced LLMs, GPT-4-Turbo.

\subsubsection{Quantifying Preferences for Properties} 

For a property \( p_i \), we calculate the degree of preference  \( P(p_i) \) as \( \sigmoid(\alpha_i) \). This value corresponds to the probability that a response is preferred over another if it only satisfies \( p_i \) better and all else equal. A higher value indicates a stronger preference, while a value less than 50\% signifies dislike.

\subsubsection{Results} 
We sort the properties by $P(p_i)$ for human and GPT-4-Turbo, and show the top/last 3 preferred ones on 9 selected groups of samples in Table \ref{tab:main-analysis-vis-short}, and the rest are shown in Table \ref{tab:main-analysis-vis-rest} in Appendix \ref{app:additional-results}.

\paragraph{Commonalities} We first observe that under different scenarios or query-specific cases, the compositions of preferences vary greatly for both humans and GPT-4-Turbo. We also find human and GPT-4-Turbo share some similarities like they have the same set of top 3 preferred properties for the \textit{Creative Writing} and \textit{Daily Tasks}, and in most cases, they both prefer responses with fewer severe errors and satisfy the explicit constraints in queries.

\begin{wraptable}[10]{R}{0.46\textwidth}
  \centering
  \small
  \caption{The average degree of preference on \textit{no severe/moderate/minor errors} across all scenarios for Human, GPT-4-Turbo, and the highest of the rest.}
    \begin{tabular}{lccc}
    \toprule
          & Minor & Moderate & Severe \\
    \midrule
    Human & 49.01&	52.45	&62.86 \\
    GPT-4-Turbo & 50.11	&58.00	&76.19 \\
    Highest of Rest & 50.54	&55.67&	65.27 \\
    \bottomrule
    \end{tabular}%
  \label{tab:error-sensitivity}%
\end{wraptable}%

\paragraph{Disparities} There are also many disagreements between them. We calculate the average degree of preference on the 3 \textit{Error Detection} features across all scenarios in Table \ref{tab:error-sensitivity} and find humans are significantly less sensitive to severe errors, and do not show a clear preference/dislike to responses with fewer moderate and minor errors. Besides, we see humans clearly dislike a model when it admits its limit in addressing the query, especially for \textit{Unsafe Query} and \textit{Communication} scenarios, indicating that human users in real settings have an urgent desire to have all their queries addressed even if they are unsafe. Humans also prefer responses that support their subjective stances (known as sycophancy), and pay little attention to how well a response corrects the mistakes or biases in queries. 

On the contrary, GPT-4-Turbo emphasizes correctness, clarity, and harmlessness more. It tends to have much larger degrees of preference for properties it prefers than humans. It also likes the responses that help clarify the unclear intent in queries and correct the mistakes for queries that are unsafe or contain mistakes/biases, which highlight the Helpfulness, Harmlessness and Honesty (HHH) goals it has been aligned to.

\subsection{How similar are the preferences of different LLMs?}
\label{sec:llm-preference-similarity}

\subsubsection{Definition of Similarity}
The similarity between preferences of two LLMs is defined as the average Pearson coefficient of the weights of their fitted Bayesian logistic regression models across all scenarios:
\begin{align}
    &\rho^s_{MN}=\frac{\sum_{i=1}^{N} (\alpha_{Mi}^s - \bar{\alpha_M^s})(\alpha_{Ni}^s - \bar{\alpha_N^s})}{\sqrt{\sum_{i=1}^{N} (\alpha_{Mi}^s - \bar{\alpha_M^s})^2} \sqrt{\sum_{i=1}^{N} (\alpha_{Ni}^s - \bar{\alpha_N^s})^2}}
    \nonumber
    \\
    &\texttt{similarity}(M,N)=\frac{1}{|S|}\sum_{s\in S}\rho^s_{MN}
    \nonumber
\end{align}
where $\alpha_M^s$ denotes the weights of the fitted Bayesian logistic regression model for an LLM (or human) $M$ on scenario $s$, $S$ is the set of all scenarios, $N=29$ is the number of defined properties.

\subsubsection{Results}
\begin{wrapfigure}[17]{R}{0.5\textwidth}
    \centering
    \includegraphics[width=0.95\linewidth]{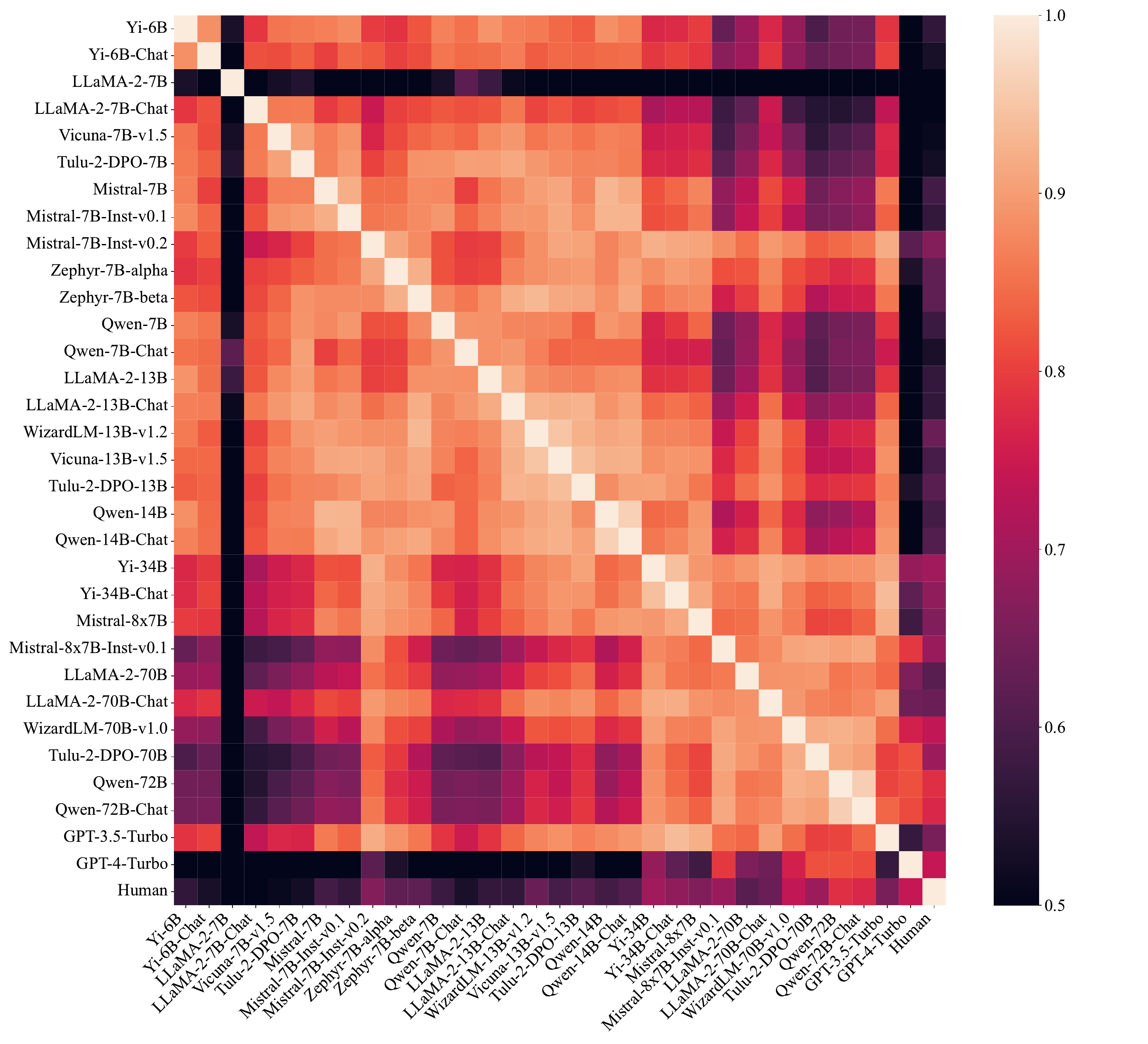}
    \caption{The similarity of preferences between different LLMs (and human). We list the LLMs in order of model size, from small to large.}
    \label{fig:preference_sim_among_models}
\end{wrapfigure}
We show the similarity in preferences between different LLMs (or human and an LLM) in Figure \ref{fig:preference_sim_among_models}. We find that the preferences of human and GPT-4-Turbo are significantly different from other LLMs, and most similar to Qwen-72B (-Chat). We also find LLaMA-2-7B is quite different from all other LLMs. This may be because it does not exhibit a consistent preference, preventing us from fitting a good model  (only 63\% accuracy, Table \ref{tab:bayeslr_acc}). 

\paragraph{Implications of LLM Size on Preferences} We divide the LLMs into two groups by their size: less than 14B (<14B) and larger than 30B (>30B). We calculate the intra- and inter-group similarities: 
\begin{align}
\text{Intra}(A) &= \frac{1}{\binom{n}{2}} \sum_{i=1}^{n-1} \sum_{j=i+1}^{n} \texttt{similarity}(a_i, a_j) \nonumber\\
\text{Inter}(A, B) &= \frac{1}{nm} \sum_{i=1}^{n} \sum_{j=1}^{m} \texttt{similarity}(a_i, b_j) \nonumber
\end{align}
where $A,B$ are groups of LLMs and $a_i,b_i$ are the components. We find the intra-group similarities (0.83 for <14B and 0.88 for >30B) are much higher than the inter-group similarity (0.74). We also divide the models by the series mentioned in \S \ref{sec:selected_models}, and find the average intra- and inter-group similarities are very close (both 0.81, a complete version in Figure \ref{fig:intra-inter-sim-series} in Appendix \ref{app:additional-results}). This further suggests that LLMs of similar sizes often have alike preferences, regardless of their training methods. 

\paragraph{Effects of Alignment on Preferences} Another question we care about is how the preference of an LLM changes after it is fine-tuned for alignment, which is calculated as the average similarity between a pretrained-only LLM and all its aligned variants. We also calculate the average log-probability difference between ``A'' and ``B'' as the first output token in collecting preference labels.
Results in Table \ref{tab:base-vs-aligned} show that the preferences tend to remain largely unchanged after fine-tuning for alignment (except for the outlier LLaMA-2-7B), but the difference in log-probability increases a lot. This can be seen as a signal that alignment does not change the tendency of LLM preference, but greatly changes the intensity of expressing it.

\begin{table}[h]
  \centering
  \small
  \setlength{\tabcolsep}{9pt}
  \caption{The difference between a pre-trained-only LLM and all its aligned (i.e., SFT, RLHF, or DPO) variants.
  }
    \begin{tabular}{lcccccccccc}
    \toprule
    \textbf{Series} & \multicolumn{3}{c}{\textbf{LLaMA-2}} & \multicolumn{3}{c}{\textbf{Qwen}} & \multicolumn{2}{c}{\textbf{Yi}} & \multicolumn{2}{c}{\textbf{Mistral}} \\
    \cmidrule(r){2-4} \cmidrule(r){5-7} \cmidrule(r){8-9} \cmidrule(r){10-11}
    \textbf{Size}  & \textbf{7B}    & \textbf{13B}   & \textbf{70B}   & \textbf{7B}    & \textbf{14B}   & \textbf{72B}   & \textbf{6B}    & \textbf{34B}   & \textbf{7B}    & \textbf{8x7B} \\
    \midrule
    \multicolumn{11}{l}{\textit{*~The average preference similarity between a base model and all its aligned variants}} \\
          & 0.52  & 0.89  & 0.89  & 0.89  & 0.96  & 0.96  & 0.88  & 0.94  & 0.88  & 0.84 \\
    \midrule
    \multicolumn{11}{l}{\textit{*~The average log-probability difference between ``A'' and ``B'' as the first output token}} \\
    pre-trained only & 0.49  & 0.47  & 0.68  & 0.76  & 0.78  & 2.24  & 0.6   & 1.98  & 0.92  & 1.01 \\
    avg. of aligned variants & 3.03  & 1.99  & 4.06  & 2.93  & 1.66  & 4.47  & 2.89  & 3.06  & 5.46  & 7.61 \\
    \bottomrule
    \end{tabular}%
  \label{tab:base-vs-aligned}%
\end{table}%

\begin{table*}
  \centering
  \small
  \caption{The results of MT-Bench (first round) and AlpacaEval 2.0 when taking strategies to inject the most and least preferred properties of different judges into the models. None in the Training-free setting means models without a system message; in the Training-based setting, it means the untrained models. $\dagger$We randomly pick a subset with 80 queries to collect human judgments.}
  \scalebox{0.95}{
    \begin{tabular}{rllllll}
    \toprule
    \multicolumn{1}{c}{\multirow{3}[2]{*}{\textbf{Model}}} & 
    \multicolumn{1}{l}{\multirow{3}[2]{*}{\textbf{Strategy}}} 
    
    & \multicolumn{2}{c}{\textbf{MT-Bench (Round 1, Rating 1-10)}} & \multicolumn{3}{c}{\textbf{AlpacaEval 2.0 (Winrate \% v.s. GPT-4-Turbo)}} \\
          \cmidrule(r){3-4} \cmidrule(r){5-7}
          &      & \multicolumn{1}{c}{\textbf{GPT-3.5-Turbo}} & \multicolumn{1}{c}{\textbf{GPT-4-Turbo}} & \multicolumn{1}{c}{\textbf{GPT-3.5-Turbo}} & \multicolumn{1}{c}{\textbf{GPT-4-Turbo}} & \multicolumn{1}{c}{\textbf{Human$\dagger$}} \\
    \midrule
    \midrule
    \multicolumn{7}{l}{\textit{*~Training-free: Setting system message to encourage models to satisfy top/last 3 preferred properties.}}\\
    \multicolumn{1}{c}{\multirow{3}[2]{*}{LLaMA-2-70B-Chat}} & None  & 7.45 & 6.21 & 42.71 & 13.87 & 35.00 \\
          & \textcolor{bettergreen}{Top 3} & 7.56 (\textcolor{bettergreen}{$\uparrow$0.11}) & 6.31 (\textcolor{bettergreen}{$\uparrow$0.10}) & 55.59 (\textcolor{bettergreen}{$\uparrow$12.88}) & 13.30 (\textcolor{betterred}{$\downarrow$0.57}) & 38.75 (\textcolor{bettergreen}{$\uparrow$3.75}) \\
          & \textcolor{betterred}{Last 3} & 7.34 (\textcolor{betterred}{$\downarrow$0.11}) & 6.05 (\textcolor{betterred}{$\downarrow$0.16}) & 39.51 (\textcolor{betterred}{$\downarrow$3.20}) & 8.77~~ (\textcolor{betterred}{$\downarrow$5.10}) & 32.50 (\textcolor{betterred}{$\downarrow$2.50}) \\
    \midrule
    \multicolumn{1}{c}{\multirow{3}[2]{*}{Qwen-72B-Chat}} & None  & 8.00 & 6.88 & 17.08 & 10.32 & 28.75 \\
          & \textcolor{bettergreen}{Top 3} & 8.16 (\textcolor{bettergreen}{$\uparrow$0.16}) & 7.08 (\textcolor{bettergreen}{$\uparrow$0.20}) & 31.55 (\textcolor{bettergreen}{$\uparrow$14.47}) & 11.03 (\textcolor{bettergreen}{$\uparrow$0.71}) & 28.75 (\textcolor{gray}{0.00}) \\
          & \textcolor{betterred}{Last 3} & 7.96 (\textcolor{betterred}{$\downarrow$0.04}) & 6.73 (\textcolor{betterred}{$\downarrow$0.15}) & 15.06 (\textcolor{betterred}{$\downarrow$2.02}) & 9.90~~ (\textcolor{betterred}{$\downarrow$0.42}) & 23.75 (\textcolor{betterred}{$\downarrow$5.00}) \\
    \midrule
    \multicolumn{1}{c}{\multirow{3}[2]{*}{GPT-3.5-Turbo}} & None  & 8.15 & 7.63 & 13.66 & 9.18 & 23.75 \\
          & \textcolor{bettergreen}{Top 3} & 8.28 (\textcolor{bettergreen}{$\uparrow$0.13}) & 7.73 (\textcolor{bettergreen}{$\uparrow$0.10}) & 42.49 (\textcolor{bettergreen}{$\uparrow$28.83}) & 11.94 (\textcolor{bettergreen}{$\uparrow$2.76}) & 25.00 (\textcolor{bettergreen}{$\uparrow$1.25}) \\
          & \textcolor{betterred}{Last 3} & 7.56 (\textcolor{betterred}{$\downarrow$0.59}) & 7.13 (\textcolor{betterred}{$\downarrow$0.50}) & 12.97 (\textcolor{betterred}{$\downarrow$0.69}) & 6.06~~ (\textcolor{betterred}{$\downarrow$3.12}) & 3.75~~ (\textcolor{betterred}{$\downarrow$20.00}) \\
    \midrule
    \multicolumn{1}{c}{\multirow{3}[2]{*}{GPT-4-Turbo}} & None  & 8.49 & 8.80 & 50.00 & 50.00 & 50.00 \\
          & \textcolor{bettergreen}{Top 3} & 8.86 (\textcolor{bettergreen}{$\uparrow$0.37}) & 8.88 (\textcolor{bettergreen}{$\uparrow$0.08}) & 81.94 (\textcolor{bettergreen}{$\uparrow$31.94}) & 50.76 (\textcolor{bettergreen}{$\uparrow$0.76}) & 50.00 (\textcolor{gray}{0.00}) \\
          & \textcolor{betterred}{Last 3} & 8.23 (\textcolor{betterred}{$\downarrow$0.26}) & 8.71 (\textcolor{betterred}{$\downarrow$0.09}) & 52.04 (\textcolor{bettergreen}{$\uparrow$2.04}) & 22.08 (\textcolor{betterred}{$\downarrow$27.92}) & 35.00 (\textcolor{betterred}{$\downarrow$15.00}) \\
    \midrule
    \midrule
    \multicolumn{7}{l}{\textit{*~Training-based: Fine-tuning the model towards/against the preferences via DPO.}}\\
    \multicolumn{1}{c}{\multirow{3}[2]{*}{Alpaca-7B}} & None  & 5.41 & 3.90 & 6.52 & 3.08 & - \\
          & \textcolor{bettergreen}{Towards} & 6.15 (\textcolor{bettergreen}{$\uparrow$0.74}) & 4.61 (\textcolor{bettergreen}{$\uparrow$0.71}) & 17.34 (\textcolor{bettergreen}{$\uparrow$10.82}) & 4.10 (\textcolor{bettergreen}{$\uparrow$1.02}) & - \\
          & \textcolor{betterred}{Against} & 4.46 (\textcolor{betterred}{$\downarrow$0.95}) & 2.88 (\textcolor{betterred}{$\downarrow$1.02}) & 4.08~~ (\textcolor{betterred}{$\downarrow$2.44}) & 1.96 (\textcolor{betterred}{$\downarrow$1.12}) & - \\
    \midrule
    \multicolumn{1}{c}{\multirow{3}[2]{*}{Alpaca-13B}} & None  & 5.46 & 3.95 & 6.15 & 2.49 & - \\
          & \textcolor{bettergreen}{Towards} & 5.60 (\textcolor{bettergreen}{$\uparrow$0.14}) & 3.98 (\textcolor{bettergreen}{$\uparrow$0.03}) & 9.20 (\textcolor{bettergreen}{$\uparrow$3.05}) & 2.93 (\textcolor{bettergreen}{$\uparrow$0.44}) & - \\
          & \textcolor{betterred}{Against} & 4.50 (\textcolor{betterred}{$\downarrow$0.96}) & 3.05 (\textcolor{betterred}{$\downarrow$0.90}) & 4.46 (\textcolor{betterred}{$\downarrow$1.69}) & 1.65 (\textcolor{betterred}{$\downarrow$0.84}) & - \\
    \bottomrule
    \end{tabular}%
  }
  \label{tab:exp-alpacaeval-mtbench}%
\end{table*}%

\subsection{Can Preference-based Evaluation be Intentionally Manipulated?}
\label{sec:manipulate-benchmarks}
As an application of dissecting the preferences, we show that results on popular benchmarks with LLM-as-a-judge can be intentionally manipulated by adapting the responses of a model to more closely align with or deliberately diverge from the identified preferences of the judge.

\subsubsection{Benchmarks} We take two benchmarks and use GPT-3.5-Turbo and GPT-4-Turbo as judges for both of them with their official evaluation prompts:
(1) 
{AlpacaEval 2.0}
~\citep{alpaca_eval} has 805 queries where each of them has a reference response generated by GPT-4-Turbo. The metric is the pairwise comparison win rate against the reference responses.
(2)
{MT-Bench}~\citep{zheng2023judging}
has 80 two-round conversations across various domains. The metric is the average response rating over all conversations. Since we focus on single-turn interaction, we only report the scores on the first rounds of all samples.

\subsubsection{Model Adaptation to Judge Preferences}

\paragraph{Training-free} We utilize four models proficient in handling system messages (LLaMA-2-70B-Chat, Qwen-72B-Chat, GPT-3.5-Turbo, GPT-4-Turbo) and configure system messages to prompt them to adhere to the judge's \textcolor{bettergreen}{\textit{Top 3}} or \textcolor{betterred}{\textit{Last 3}}  preferred properties in the scenario of a query. In this setting, we also sample a subset of AlpacaEval 2.0 with 80 queries and require two authors as human judges.

\paragraph{Training-based} We use the fitted Bayesian logistic regression models for our target judges to annotate the preference labels on our collected dataset. We exclude samples where the final preference probability is within $50\pm15\%$ to emphasize the preferences. We then use the remaining data (4,022 for GPT-3.5-Turbo and 3,991 for GPT-4-Turbo) to train 2 Alpaca models\footnote{https://huggingface.co/chavinlo/alpaca-7b (13b)} with 
DPO {\color{bettergreen}\textit{towards}} the preferences of judges
(GPT-3.5-Turbo and GPT-4-Turbo, respectively). 
For training {\color{betterred}\textit{against}} preferences, we simply invert labels from the regression models. We set batch size as 64, learning rate as 1e-5 with a cosine scheduler, and train three epochs.

\subsubsection{Results}
The results are presented in Table \ref{tab:exp-alpacaeval-mtbench}. Generally, we find that in both training-free and training-based settings, adapting model responses to align with or diverge from judge preferences results in corresponding improvements or reductions in scores. Compared to GPT-4-Turbo, the effect of adaptation is more noticeable when targeting GPT-3.5-Turbo, possibly due to its 
less robust inferential abilities.
While the effect is somewhat less pronounced with human judges, the manipulation remains effective, demonstrating its generality.

It is also essential to highlight that the prompts for LLM-as-a-judge in both benchmarks differ notably from those we employed for collecting preferences. These prompts included extra instructions, and MT-Bench even uses a single-response rating system. Despite the differences, our approach, which relies on pairwise comparisons to gather preferences, shows substantial and predictable effectiveness. This outcome underscores the robustness of our method in preference analysis, demonstrating its reliability even under varied conditions.

\section{Conclusion}
In this work, we conduct a thorough analysis to dissect how the preferences of human and LLMs can be quantitatively decomposed into different properties. We find that humans prefer responses that can directly address their queries but are less sensitive to errors in responses, while advanced LLMs like GPT-4-Turbo emphasize more on correctness, clarity, and harmlessness. We then find that model size can be a distinguishing factor in the preferences where LLMs of similar sizes share similar preferences, and fine-tuning for alignment does not bring about significant change to LLM preferences. As an application of our analysis, we also show the results of current benchmarks with LLM-as-a-judge can be intentionally manipulated, indicating the vulnerability of preference-based evaluation. Last but not least, we publicly release all the collected resources to facilitate future research.

\section{Limitation and Future Work}
To obtain a binary preference label without positional bias, our method is not applicable to proprietary LLMs like Gemini \citep{team2023gemini} or Claude \citep{anthropic2023claude}, which do not return log-probabilities. Additionally, our use of Bayesian logistic regression for preference dissection assumes that all predefined properties independently influence the final preference, possibly missing complex interactions between them. Designing predefined properties and corresponding prompts for GPT-4-Turbo is still challenging and laborious. A future improvement could be the automatic discovery of these properties, avoiding the trial and error in creating effective prompts for data annotation. This work focuses only on single-turn conversations, leaving the extension to more complex, multi-turn interactions for future research.

\bibliography{custom}
\bibliographystyle{acl_natbib}

\appendix
\newpage

\section{Detailed Properties and Prompts for GPT-4-Turbo Annotation}
\label{app:prompt}
The prompt for annotating the \textit{Basic} properties is in Figure \ref{fig:basic-properties-prompt}. The detailed descriptions for the basic properties are in Figure \ref{fig:basic-properties-q-free} and \ref{fig:basic-properties-q-aware}.
The prompt for the first-round annotation, i.e. the prerequisite questions of the \textit{Query-specfic} properties is in Figure \ref{fig:q-check-prompt}. The prompt for the second-round annotation for query-specific properties is in Figure~\ref{fig:q-aware-twice-prompt}, and the detailed contents for the placeholders \texttt{\{questions str\}} and \texttt{\{output format str\}} in it are in Figure~\ref{fig:q-aware-twice-questions} and \ref{fig:q-aware-twice-outputs} respectively.
The prompt for \textit{Error Detection} is shown in Figure~\ref{fig:accuracy-prompt}.

\section{Defined Scenarios}
\label{app:scenario_group}
We show how we merge the scenarios from the classifier in \citet{li2023generative} into our newly-defined 10 scenarios in Table \ref{tab:scenario-merge}.

\begin{table}[h]
  \centering
  \small
  \caption{The 10 new scenarios and the mapping from scenarios defined in \citet{li2023generative} to ours.}
    \begin{tabular}{lp{12cm}}
\toprule
\textbf{Our Scenarios} & \textbf{Scenarios in \citet{li2023generative}} \\
\midrule
\multirow{1}[0]{*}{Exam Questions} & math-reasoning, solving-exam-question-with-math, solving-exam-question-without-math \\
\rowcolor{gray!15}\multirow{2}[0]{*}{Code} & code-simplification, code-generation, explaining-code, code-correction-rewriting, code-to-code-translation \\
\multirow{3}[0]{*}{Creative Writing} & writing-song-lyrics, writing-social-media-post, writing-blog-post, writing-personal-essay, creative-writing, writing-advertisement, writing-marketing-materials, writing-presentation-script, counterfactual \\
\rowcolor{gray!15}\multirow{3}[0]{*}{Functional Writing} & writing-product-description, writing-job-application, writing-news-article, writing-biography, writing-email, writing-legal-document, writing-technical-document, writing-scientific-paper, functional-writing, writing-cooking-recipe \\
Communication & value-judgement, chitchat \\
\rowcolor{gray!15}Knowledge-aware & open-question, explaining-general, verifying-fact \\
Advice & asking-how-to-question, seeking-advice \\
\rowcolor{gray!15}Daily Tasks & analyzing-general, roleplay, planning, recommendation, brainstorming \\
\multirow{5}[0]{*}{NLP Tasks} & ranking, text-to-text-translation, classification-identification, title-generation, question-generation, reading-comprehension, keywords-extraction, information-extraction, topic-modeling, data-analysis, post-summarization, text-summarization, note-summarization, text-simplification, language-polishing, instructional-rewriting, text-correction, paraphrasing \\
\rowcolor{gray!15}Others & default \\
\bottomrule
\end{tabular}%
  \label{tab:scenario-merge}%
\end{table}%

\begin{wrapfigure}[19]{R}{0.5\textwidth}
    \centering
    \includegraphics[width=0.95\linewidth]{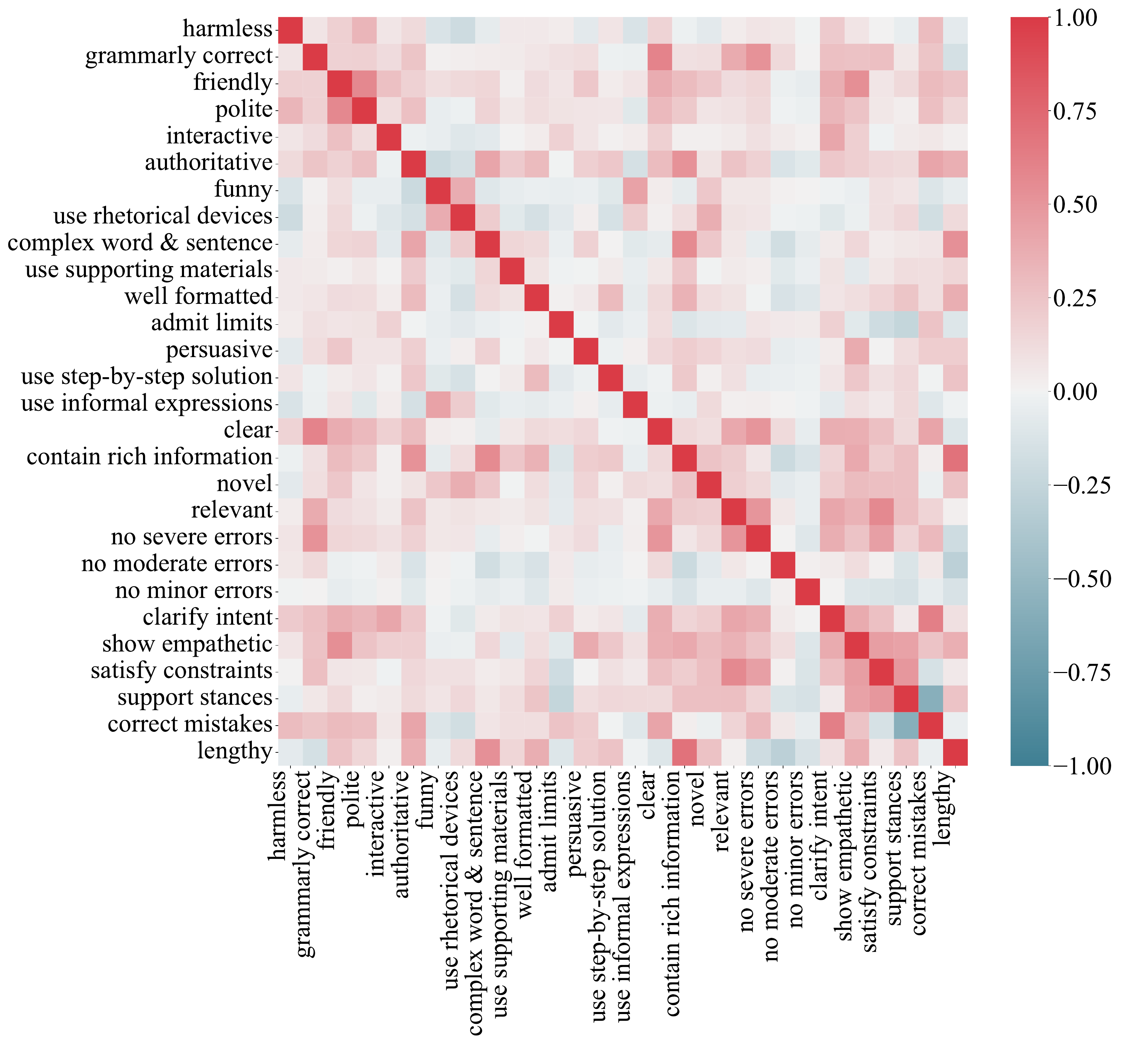}
    \caption{Property correlation in the annotated data.}
    \label{fig:data-stat-corr}
\end{wrapfigure}

\section{Details of Selected Open-Source LLMs}
\label{app:details-of-open-source-llms}
We select 30 recent open-source LLMs from different series: LLaMA-2 \citep{touvron2023llama2}, Mistral \citep{jiang2023mistral,jiang2023mixtral}, Vicuna \citep{vicuna2023}, WizardLM \citep{xu2023wizardlm}, Tulu-2 \citep{ivison2023tulu2}, Yi \citep{01aiyi}, Zephyr \citep{tunstall2023zephyr}, Qwen \citep{bai2023qwen}. The details of all selected open-source LLMs are listed in Table \ref{tab:opensource-llms-details}, including the base model they are trained on, and whether they are only pre-trained or aligned via Supervised Fine-tuning (SFT), RLHF, or DPO.

\section{Annotation Example}
\label{app:annotation-of-a-specific-sample}
We present an example in Table \ref{tab:shap-example-details} to illustrate the contents of a fully annotated sample in our collected dataset.

\section{More Statistics of Collected Dataset}
\label{app:data-stat-more}
We show more statistics of the collected dataset, including the average score/count for each property in Table \ref{tab:data-stat-mean} and how different properties correlate with each other in Figure \ref{fig:data-stat-corr}.

\begin{minipage}[t]{.45\textwidth} 
    \captionof{table}{The detailed information of the selected open-source LLMs for preferences collection. We list the base models they are trained on, as well as whether they are only pre-trained or aligned via techniques like Instruction Tuning (SFT), RLHF, or DPO.} 
    \scriptsize
    \centering
    \begin{tabular}{lllc}
    \toprule
    \textbf{Series} & \textbf{Version} & \textbf{Base Model} & \textbf{Aligned} \\
    \midrule
    \multirow{6}[1]{*}{LLaMA-2} & 7B    & LLaMA-2-7B & \ding{55} \\
          & 7B-Chat & LLaMA-2-7B & \ding{51} \\
          & 13B   & LLaMA-2-13B & \ding{55} \\
          & 13B-Chat & LLaMA-2-13B & \ding{51} \\
          & 70B   & LLaMA-2-70B & \ding{55} \\
          & 70B-Chat & LLaMA-2-70B & \ding{51} \\
    \midrule
    \multirow{2}[0]{*}{Vicuna} & 7B-v1.5 & LLaMA-2-7B & \ding{51} \\
          & 13B-v1.5 & LLaMA-2-13B & \ding{51} \\
    \midrule
    \multirow{2}[0]{*}{WizardLM} & 13B-v1.2 & LLaMA-2-13B & \ding{51} \\
          & 70B-v1.0 & LLaMA-2-70B & \ding{51} \\
    \midrule
    \multirow{5}[0]{*}{Mistral} & 7B    & Mistral-7B & \ding{55} \\
          & 7B-Inst-v0.1 & Mistral-7B & \ding{51} \\
          & 7B-Inst-v0.2 & Mistral-7B & \ding{51} \\
          & 8x7B  & Mistral-8x7B & \ding{55} \\
          & 8x7B-Inst-v0.1 & Mistral-8x7B & \ding{51} \\
    \midrule
    \multirow{3}[0]{*}{Tulu-2} & 7B-DPO & LLaMA-2-7B & \ding{51} \\
          & 13B-DPO & LLaMA-2-13B & \ding{51} \\
          & 70B-DPO & LLaMA-2-70B & \ding{51} \\
    \midrule
    \multirow{4}[0]{*}{Yi} & 6B    & Yi-6B & \ding{55} \\
          & 6B-Chat & Yi-6B & \ding{51} \\
          & 34B   & Yi-34B & \ding{55} \\
          & 34B-Chat & Yi-34B & \ding{51} \\
    \midrule
    \multirow{2}[0]{*}{Zephyr} & 7B-Alpha & Mistral-7B & \ding{51} \\
          & 7B-Beta & Mistral-7B & \ding{51} \\
    \midrule
    \multirow{6}[1]{*}{Qwen} & 7B    & Qwen-7B & \ding{55} \\
          & 7B-Chat & Qwen-7B & \ding{51} \\
          & 14B   & Qwen-14B & \ding{55} \\
          & 14B-Chat & Qwen-14B & \ding{51} \\
          & 72B   & Qwen-72B & \ding{55} \\
          & 72B-Chat & Qwen-72B & \ding{51} \\
    \bottomrule
    \end{tabular}%
    \label{tab:opensource-llms-details}%
\end{minipage}%
\hfill 
\begin{minipage}[t]{.45\textwidth} 
    \captionof{table}{Mean Score/Count for each property in collected data. $\dagger$The average scores of 5 query-specific properties are calculated only on samples where the queries met specific prerequisites.} 
    \small
    \centering
    \begin{tabular}{lc}
    \toprule
    Property & Mean Score \\
    \midrule
    harmless       &      2.90\\
    grammarly correct        &      2.70\\
    friendly           &      1.79\\
    polite             &      2.78\\
    interactive        &      0.22\\
    authoritative      &      1.67\\
    funny              &      0.08\\
    use rhetorical devices      &      0.16\\
    complex word \& sentence &   0.89\\
    use supporting materials  &   0.13\\
    well formatted        &   1.26\\
    admit limits          &   0.17\\
    persuasive              &   0.27\\
    step-by-step          &   0.37\\
    use informal expressions  &   0.04\\
    clear                 &   2.54\\
    contain rich information         &   1.74\\
    novel                 &   0.47\\
    relevant             &   2.45\\
    clarify intent$\dagger$        &   1.33\\ 
    show empathetic$\dagger$       &   1.48\\
    satisfy constraints$\dagger$   &   2.01\\
    support stances$\dagger$       &   2.28\\
    correct mistakes$\dagger$      &   1.08\\
    \midrule
    Property & Mean Count \\
    \midrule
    severe errors          &  0.59  \\
    moderate errors        &  0.61  \\
    minor errors           &  0.23  \\
    length                & 164.52  \\
    \bottomrule
    \end{tabular}%
    \label{tab:data-stat-mean}%
\end{minipage}

\begin{table}
  \centering
  \small
  \setlength{\tabcolsep}{4pt}
  \caption{An example of a fully annotated sample in our collected dataset. For each property, we show the annotation results for of the two responses: first \texttt{Response A}, then \texttt{Response B}.}
  \scalebox{0.95}{
    \begin{tabularx}{\textwidth}{XXXXXXXXX}
    \toprule
    \multicolumn{9}{l}{\textbf{Query (Scenario: Daily Tasks)}} \\
    \midrule
    \multicolumn{9}{p{0.99\textwidth}}{give me an itinerary for a day in paris on june 11th saturday from 12pm to 10pm. no museums or other things that take too long. include lunch and dinner (vegan options preferred)} \\
    \midrule
    \multicolumn{9}{l}{\textbf{Response A (102 words), Preferred by User}} \\
    \midrule
    \multicolumn{9}{p{0.99\textwidth}}{12pm: Have lunch at a vegan restaurant in the Latin Quarter, such as Sage. 

2pm: Take a stroll down the Rue Mouffetard, a bustling market street. 

4pm: Visit Notre Dame Cathedral and take a look around the grounds. 

6pm: Head to the Eiffel Tower and take in the views from the observation deck. 

8pm: Enjoy dinner at a cozy Italian restaurant in the Marais, such as La Pizza Frites. 

10pm: Wrap up the day with a stroll along the Seine, taking in the beautiful city lights.} \\
    \midrule
    \multicolumn{9}{l}{\textbf{Response B (346 words)}} \\
    \midrule
    \multicolumn{9}{p{0.99\textwidth}}{Sure, here is a suggested itinerary for a day in Paris on June 11th, 2021, that includes lunch and dinner options and keeps the activities relatively short:

12:00pm: Arrive at the Sainte-Chapelle, a beautiful Gothic chapel that was built in the 13th century. The chapel is famous for its stunning stained glass windows, which are considered some of the most beautiful in Europe.

1:00pm: Walk along the River Seine and take a Seine River cruise. This allows you to see the city from a different perspective and also visit some of the city's most famous landmarks such as the Eiffel Tower, the Notre-Dame and the Île de la Cité.

2:30pm: Have lunch at Chez L'Ami Jean, a popular vegan bakery located in the Montmartre neighborhood. The bakery offers a variety of vegan pastries and sandwiches.

3:30pm: Visit the Arc de Triomphe, a famous landmark that was built in the early 19th century. The arc is located at the top of the Champs-Élysées, one of the most famous streets in Paris.

4:30pm: Walk to the Musée du quai Branly - Jacques Chirac, a museum that showcases the art and culture of Africa, Oceania, and the Americas. The museum's stunning architecture and unique collections make for an interesting visit.

6:30pm: Have dinner at Le Comptoir du Relais, a vegan bakery located in the Montmartre neighborhood. The bakery offers a variety of vegan pastries and sandwiches.

7:30pm: Visit the Eiffel Tower, a famous landmark that was built in the late 19th century. The tower offers stunning views of the city and is a must-see for any first-time visitor to Paris.

9:30pm: End the evening with a casual dinner at Chez L'Ami Jean, a popular vegan bakery located in the Montmartre neighborhood. The bakery offers a variety of vegan pastries and sandwiches.

Note: This itinerary is just} \\
    \midrule
    \multicolumn{9}{l}{\textbf{Annotations For Basic Features (ratings, higher value means satisfying the property better)}} \\
    \midrule
        \multicolumn{1}{p{0.25\textwidth}}{harmless} & \multicolumn{1}{r}{3} & \multicolumn{1}{r|}{3} & 
        \multicolumn{1}{p{0.25\textwidth}}{grammarly correct} & \multicolumn{1}{r}{3} & \multicolumn{1}{r|}{2} & 
        \multicolumn{1}{p{0.24\textwidth}}{friendly} & \multicolumn{1}{r}{2} & \multicolumn{1}{r}{2} \\
    
        \multicolumn{1}{p{0.25\textwidth}}{polite} & \multicolumn{1}{r}{3} & \multicolumn{1}{r|}{3} & 
        \multicolumn{1}{p{0.25\textwidth}}{interactive} & \multicolumn{1}{r}{0} & \multicolumn{1}{r|}{0} & 
        \multicolumn{1}{p{0.24\textwidth}}{authoritative} & \multicolumn{1}{r}{2} & \multicolumn{1}{r}{2} \\
        
        \multicolumn{1}{p{0.25\textwidth}}{funny} & \multicolumn{1}{r}{0} & \multicolumn{1}{r|}{0} & 
        \multicolumn{1}{p{0.25\textwidth}}{use rhetorical devices} & \multicolumn{1}{r}{0} & \multicolumn{1}{r|}{0} & 
        \multicolumn{1}{p{0.24\textwidth}}{complex word \& sentence} & \multicolumn{1}{r}{1} & \multicolumn{1}{r}{1} \\
    
        \multicolumn{1}{p{0.25\textwidth}}{use supporting materials} & \multicolumn{1}{r}{0} & \multicolumn{1}{r|}{0} & 
        \multicolumn{1}{p{0.25\textwidth}}{well formatted} & \multicolumn{1}{r}{2} & \multicolumn{1}{r|}{2} & 
        \multicolumn{1}{p{0.24\textwidth}}{admit limits} & \multicolumn{1}{r}{0} & \multicolumn{1}{r}{0} \\
        
        \multicolumn{1}{p{0.25\textwidth}}{persuasive} & \multicolumn{1}{r}{0} & \multicolumn{1}{r|}{0} & 
        \multicolumn{1}{p{0.25\textwidth}}{step-by-step} & \multicolumn{1}{r}{0} & \multicolumn{1}{r|}{0} & 
        \multicolumn{1}{p{0.24\textwidth}}{use informal expressions} & \multicolumn{1}{r}{0} & \multicolumn{1}{r}{0} \\
    
        \multicolumn{1}{p{0.25\textwidth}}{non-repetitive} & \multicolumn{1}{r}{3} & \multicolumn{1}{r|}{2} & 
        \multicolumn{1}{p{0.25\textwidth}}{clear} & \multicolumn{1}{r}{3} & \multicolumn{1}{r|}{2} & 
        \multicolumn{1}{p{0.24\textwidth}}{relevant} & \multicolumn{1}{r}{3} & \multicolumn{1}{r}{1} \\
        
        \multicolumn{1}{p{0.25\textwidth}}{novel} & \multicolumn{1}{r}{1} & \multicolumn{1}{r|}{1} & 
        \multicolumn{1}{p{0.25\textwidth}}{contain rich information} & \multicolumn{1}{r}{2} & \multicolumn{1}{r|}{3} &   
        \multicolumn{1}{p{0.24\textwidth}}{} &    {}   & {} \\
    
    \midrule
    \multicolumn{9}{l}{\textbf{Annotations for Query-Specific Features, Both Rounds.}} \\
    \midrule[0.1pt]
    \multicolumn{1}{l@{}}{Clear Intent} & \multicolumn{2}{l|}{Yes} & 
    \multicolumn{1}{l@{}}{Explicitly express feeling} & \multicolumn{2}{l|}{No} & 
    \multicolumn{1}{l}{Mistakes or Bias} & \multicolumn{2}{r}{None} \\
    \multicolumn{1}{l@{}}{Explicit Constraints} & \multicolumn{2}{l|}{5} & \multicolumn{1}{l@{}}{Subjective Stances} & \multicolumn{2}{l|}{2} &       &       &  \\
    \midrule
    \multicolumn{9}{l}{\texttt{5 Constraints}\hspace{19.3em}\texttt{2 Stances}} \\
    
    \multicolumn{9}{l}{Itinerary for a day in Paris on June 11th, Saturday:\hspace{0.6em}3\quad3\hspace{3em}Preference for a day without visiting museums or lengthy} \\
    
    \multicolumn{9}{l}{Time frame from 12pm to 10pm:\hspace{7.6em}3\quad2\hspace{3em}activities:\hspace{4.4em}Strongly supported\quad Weakly opposed} \\
    
    \multicolumn{9}{l}{No museums or other activities that take too long:\hspace{0.9em}3\quad1\hspace{3em}Preference for vegan meal options:}\\
    
    \multicolumn{9}{l}{Include lunch and dinner:\hspace{10.6em}3\quad3\hspace{11.2em}Strongly supported\quad Weakly supported}\\
    
    \multicolumn{9}{l}{Vegan options preferred for meals:\hspace{7em}3\quad2}\\
    \midrule
    
    \multicolumn{9}{l}{\textbf{Annotations for Error Detection}} \\
    \midrule
    \multicolumn{9}{p{0.99\textwidth}}{\texttt{Response A:} 
    
    Moderate (\textit{Information contradiction to the query}) - La Pizza Frites is not known as a vegan restaurant, which contradicts the user's preference for vegan dinner options.} \\
    \midrule[0.1pt]
    \multicolumn{9}{p{0.99\textwidth}}{\texttt{Response B:} 

    Severe (\textit{Factual error}) - Le Comptoir du Relais is not a vegan bakery, and it is incorrectly listed as such.
    
    Moderate (\textit{Information contradiction to the query}) - Sainte-Chapelle visit contradicts the 'no museums or other things that take too long' request.
    
    Moderate (\textit{Information contradiction to the query}) - Chez L'Ami Jean is not a vegan bakery, and it is listed for both lunch and dinner, which contradicts the vegan preference.
    
    Moderate (\textit{Information contradiction to the query}) - Musée du quai Branly - Jacques Chirac visit contradicts the 'no museums or other things that take too long' request.
    
    Moderate (\textit{Information contradiction to the query}) - The itinerary suggests ending the evening with a casual dinner at Chez L'Ami Jean after already having dinner there at 6:30pm.
    
    Minor (\textit{Information contradiction to the query}) - The date June 11th, 2021, is incorrect as the query asks for an itinerary for June 11th without specifying a year.}\\
    \bottomrule
    \end{tabularx}%
  }
  \label{tab:shap-example-details}%
\end{table}%

\section{Converting Annotations of Query-Specific Properties to Ratings}
\label{app:query-specific-rating}
For \textit{clarify intent} and \textit{show empathetic}, we directly annotate a score of 0,1,2,3 to measure how much a response matches the property.

For \textit{satisfy constraints}, we annotate 0,1,2,3 for each identified constraint on how each of them is satisfied and take the average value over all constraints as the final rating.

For \textit{support stances}, we use this mapping to convert the collected labels to scores: \texttt{\{strongly supported: 3, weakly supported: 2.25, neutral: 1.5, weakly opposed: 0.75, strongly opposed: 0\}}. The final rating is the average of these scores across all stances.

For \textit{correct mistakes}, we use this mapping to convert the collected labels to scores: \texttt{\{pointed out and corrected: 3, corrected without being pointed out: 2, pointed out but not corrected: 1, neither pointed out nor corrected: 0\}}. The final rating is the average of these scores across all mistakes/biases.

\section{Detailed Statistics of Annotation Quality Check by Human}
\label{app:annotation-quality-check}
We show the detailed statistics of the quality check of the 50 randomly selected annotated samples by human evaluators in Table \ref{tab:stat-annotation-quality-check}, where we ask the evaluators to choose whether they agree with each annotation item or not in a sample. These annotation items include the ratings of \textit{Basic} properties, the answers to the prerequisite questions and the optional second-round annotation result for \textit{Query-Specific} properties, and the detected errors for \textit{Error Detection}.

\begin{table*}
  \centering
  \small
  \setlength{\tabcolsep}{3.5pt}
  \caption{The agreement rate of human evaluators on the annotation results given by GPT-4-Turbo on 50 randomly selected samples from the full dataset. $x/y$ denotes in $y$ items (they can be samples, responses, or detected entities), $x$ of them are agreed upon by human evaluators. In \textit{Query-Specific} and \textit{Error Detection} properties, the annotation of a response is agreed only when all detected entities in this response are agreed.}
    \begin{tabular}{llllll}
    \toprule
    \multicolumn{6}{c}{\textbf{Basic Properties (response-level)}} \\
    \midrule
    friendly & \multicolumn{1}{l|}{95/100} & persuasive & \multicolumn{1}{l|}{93/100} & clear & 93/100 \\
    relevant & \multicolumn{1}{l|}{83/100} & admit limits & \multicolumn{1}{l|}{95/100} & novel & 97/100 \\
    step-by-step & \multicolumn{1}{l|}{93/100} & authoritative & \multicolumn{1}{l|}{85/100} & polite & 88/100 \\
    use rhetorical devices & \multicolumn{1}{l|}{98/100} & non-repetitive & \multicolumn{1}{l|}{96/100} & funny & 98/100 \\
    use supporting materials & \multicolumn{1}{l|}{95/100} & well formatted & \multicolumn{1}{l|}{87/100} & harmless & 94/100 \\
    use informal expressions & \multicolumn{1}{l|}{98/100} & contain rich info & \multicolumn{1}{l|}{94/100} & interactive & 93/100 \\
    complex word  \& sentence & \multicolumn{1}{l|}{93/100} & grammarly correct & \multicolumn{1}{l|}{94/100} &       &  \\
    \midrule
    \multicolumn{6}{c}{\textbf{Query-specific Prerequisite (sample-level)}} \\
    \midrule
    clear intent & \multicolumn{1}{l|}{48/50} & contain explicit constraints & \multicolumn{1}{l|}{44/50} & express feelings & 48/50 \\
    show explicit subjective stances & \multicolumn{1}{l|}{48/50} & contain mistakes or bias & \multicolumn{1}{l|}{49/50} &       &  \\
    \midrule
    \multicolumn{6}{c}{\textbf{Query-specific Properties}} \\
    \midrule
    satisfy constraints (response-level) & \multicolumn{1}{l|}{21/26} & correct mistakes (response-level) & \multicolumn{1}{l|}{4/4} & clarify intent & 5/6 \\
    satisfy constraints (constraint-level) & \multicolumn{1}{l|}{47/56} & correct mistakes (mistake-level) & \multicolumn{1}{l|}{4/4} & show empathetic & 0/0 \\
    support stances (response-level) & \multicolumn{1}{l|}{5/6} & support stances (stance-level) & \multicolumn{1}{l|}{9/10} &       &  \\
    \midrule
    \multicolumn{6}{c}{\textbf{Error Detection}} \\
    \midrule
    completeness (response-level) & \multicolumn{1}{l|}{90/100}  & agreement (response-level) & \multicolumn{1}{l|}{75/100} & agreement (error-level) & 160/188 \\
    \bottomrule
    \end{tabular}%
  \label{tab:stat-annotation-quality-check}%
\end{table*}%

\section{Comparison Strategy}
\label{app:comparison-strategy}
For most of the properties, we mark the comparison feature as +1/-1/0 if the rating of response A is larger/smaller than or equal to response B. For properties of \textit{Error Detection} of different severity (i.e. \textit{no severe/moderate/minor errors}), we mark the comparison feature as +1/-1/0 if response A contains less/more/equal number of errors than response B. We also have two special cases: For \textit{relevant}, we only mark a +1/-1 when one response has a rating of 0 and the other >0, otherwise we mark it as 0; for \textit{lengthy}, we mark +1/-1 when the word counts of the two responses have a significant difference, i.e., the shorter one has fewer than 70\% words of the longer one, otherwise we mark it as 0.

\section{Details of Fitting Bayesian Logistic Regression}
\label{app:bayes-lr}
We perform approximate Bayesian inference with the No-U-Turn Sampler \citep{hoffman2014no} with Hamiltonian Monte Carlo \citep{neal2011mcmc} implemented using \texttt{numpyro} \citep{phan2019composable}, collecting 6,000 posterior samples across four independent Markov Chain Monte Carlo (MCMC) chains (each chain contains 500 warmup samples and 1,500 collected samples). The scale $b=0.1$ of the Laplace prior is determined using the remaining part as the validation set in each iteration of our 10-fold aggregation.

\section{Accuracy of the Fitted Bayesian Logistic Models}

\label{app:bayeslr_acc}
We report the prediction accuracy of the fitted Bayesian Logistic models for a preference $D_j$:
$$
\texttt{acc}(D_j) = \frac{1}{|S|}\sum_{s\in S}\texttt{acc}(D_j^s)
$$
, where $D_j$ is the preference dataset of a judge $j$, $D_j^s$ is the subset of it with scenario as $s$, and $S$ is the set of all scenarios. The results are in Table \ref{tab:bayeslr_acc}.

\begin{table}[h]
    \centering
  \small
  \caption{Average prediction accuracy of all fitted Bayesian logistic models for a preference dataset $D_j$.}
    \begin{tabular}{lclclc}
    \toprule
    \textbf{Judge} & \textbf{Accuracy} & \textbf{Judge} & \textbf{Accuracy} &\textbf{Judge} & \textbf{Accuracy} \\
    \midrule
    Yi-6B  & 75.81 &Yi-6B-Chat  & 77.32 &    LLaMA-2-7B  & 63.16 \\
    LLaMA-2-7B-Chat  & 75.51 &    Vicuna-7B-v1.5  & 76.97 &    Tulu-2-DPO-7B  & 77.89 \\
    Mistral-7B  & 82.26 &    Mistral-7B-Inst-v0.1  & 80.24 &    Mistral-7B-Inst-v0.2  & 81.61 \\
    Zephyr-7B-alpha  & 82.50 &    Zephyr-7B-beta  & 80.90 &    Qwen-7B  & 76.93 \\
    Qwen-7B-Chat  & 73.75 &    LLaMA-2-13B  & 76.06 &    LLaMA-2-13B-Chat  & 78.51 \\
    WizardLM-13B-v1.2  & 80.86 &    Vicuna-13B-v1.5  & 81.03 &    Tulu-2-DPO-13B  & 80.69 \\
    Qwen-14B  & 80.69 &    Qwen-14B-Chat  & 81.30 &    Yi-34B  & 83.15 \\
    Yi-34B-Chat  & 83.42 &    Mistral-8x7B  & 82.78 &    Mistral-8x7B-Inst-v0.1  & 83.44 \\
    LLaMA-2-70B  & 82.03 &    LLaMA-2-70B-Chat  & 82.38 &    WizardLM-70B-v1.0  & 83.27 \\
    Tulu-2-DPO-70B  & 83.25 &    Qwen-72B  & 83.94 &    Qwen-72B-Chat  & 84.25 \\
    GPT-3.5-Turbo  & 84.15 &    GPT-4-Turbo  & 86.79 &    Human  & 78.12 \\
    \bottomrule
    \end{tabular}%
  \label{tab:bayeslr_acc}%
\end{table}

\begin{table}[h]
  \centering
  \scriptsize
  \setlength{\tabcolsep}{0pt}
  \caption{The top/last 3 preferred properties of Human and GPT-4-Turbo on groups of samples excluded in Table \ref{tab:main-analysis-vis-short}.}
    \begin{tabular}{cccc}
    \toprule
    \multirow{2}{*}{\parbox[c]{1cm}{\textcolor{myskyblue}{Scenario-Wise}}}     & \multirow{2}[1]{*}{Code} & \multirow{2}[1]{*}{Exam Questions}  & \multirow{2}[1]{*}{Functional Writing} \\ \\
    \midrule
    \multirow{5}{*}{\parbox[c]{1cm}{Human}} & \multirow{5}[1]{*}{\includegraphics[scale=0.395]{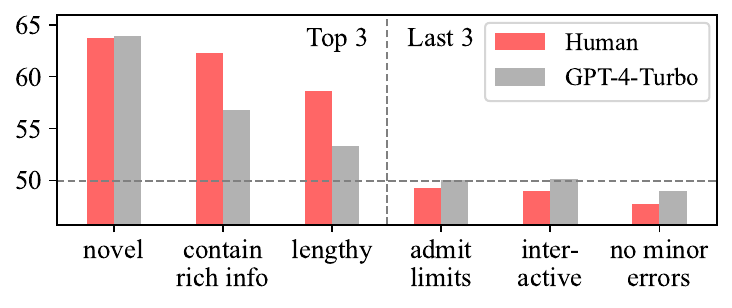}} & \multirow{5}[1]{*}{\includegraphics[scale=0.395]{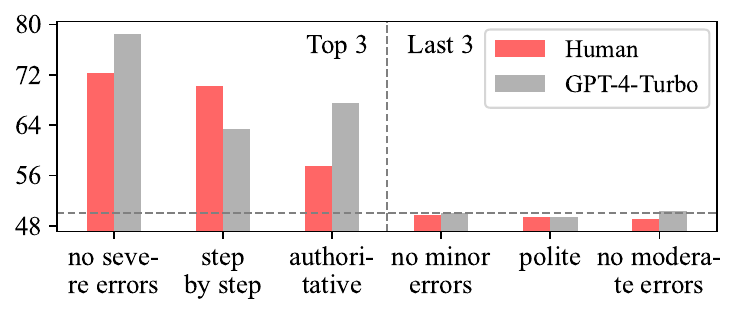}}  & \multirow{5}[1]{*}{\includegraphics[scale=0.395]{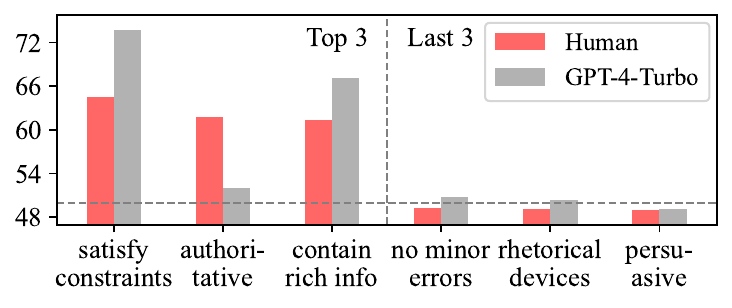}} \\
    \\ \\ \\ \\ \rule{0pt}{5mm} \\ 
    \multirow{5}{*}{\parbox[c]{1cm}{GPT-4-Turbo}}& \multirow{5}[0]{*}{\includegraphics[scale=0.395]{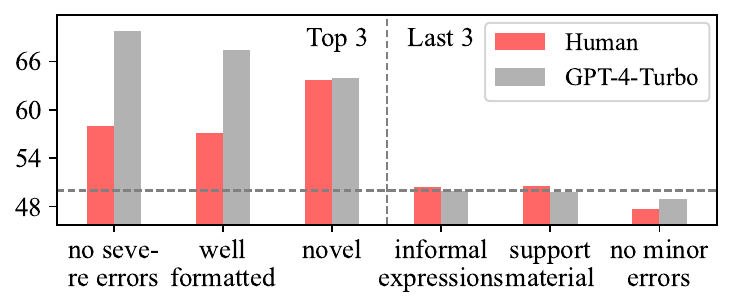}} & \multirow{5}[0]{*}{\includegraphics[scale=0.395]{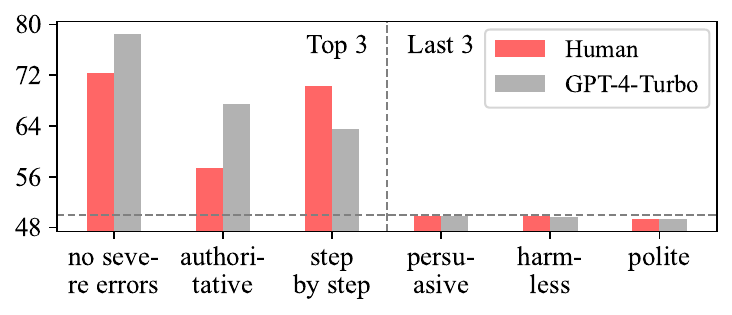}}  & \multirow{5}[0]{*}{\includegraphics[scale=0.395]{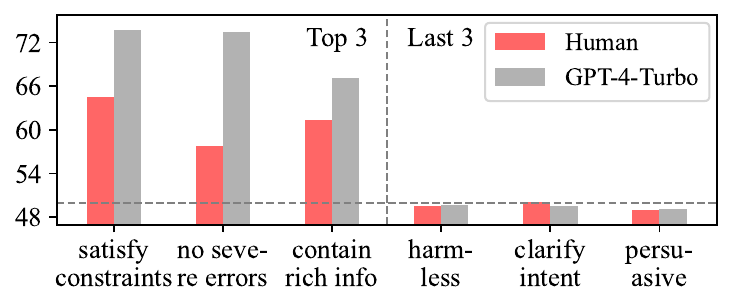}} \\
    \\ \\ \\ \\ \rule{0pt}{5mm} \\ 
    \midrule
    \multirow{2}{*}{\parbox[c]{1cm}{\textcolor{myskyblue}{Scenario-Wise}}}     & \multirow{2}[1]{*}{NLP Tasks} & \multirow{2}[1]{*}{Others}  & \\ \\
    \midrule
    \multirow{5}{*}{\parbox[c]{1cm}{Human}} & \multirow{5}[1]{*}{\includegraphics[scale=0.395]{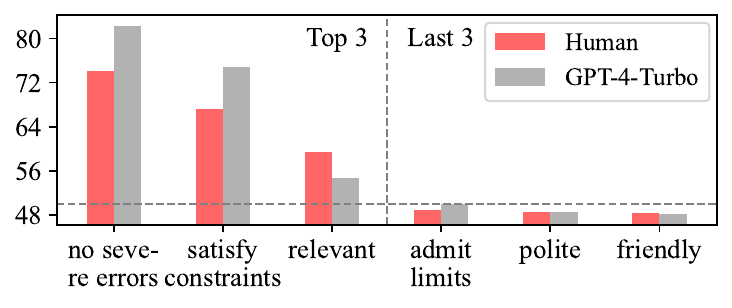}}   & \multirow{5}[1]{*}{\includegraphics[scale=0.395]{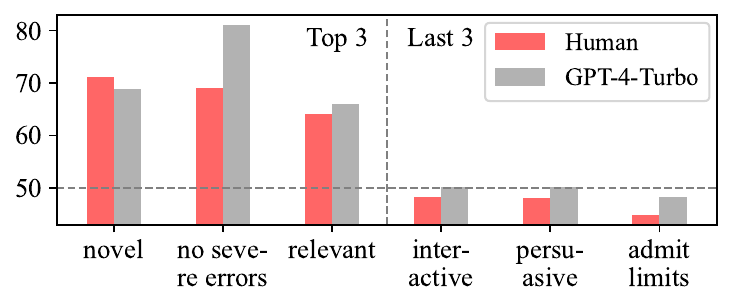}} & \\
    \\ \\ \\ \\ \rule{0pt}{5mm} \\ 
    \multirow{5}{*}{\parbox[c]{1cm}{GPT-4-Turbo}}& \multirow{5}[1]{*}{\includegraphics[scale=0.395]{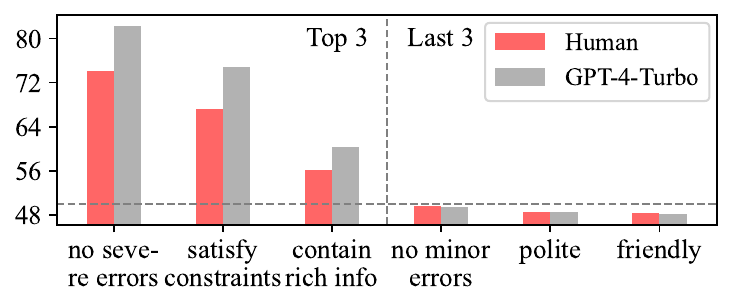}}  & \multirow{5}[1]{*}{\includegraphics[scale=0.395]{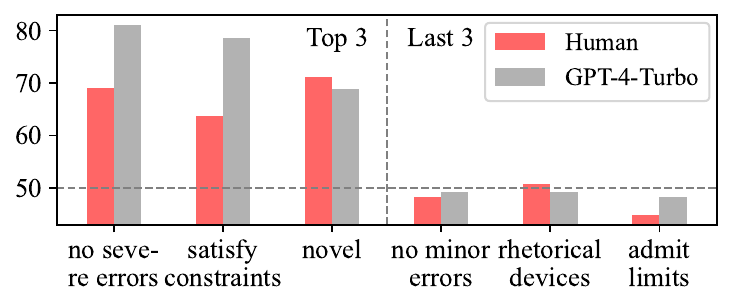}} &\\
    \\ \\ \\ \\ \rule{0pt}{5mm} \\ 
    \midrule
    \multirow{2}{*}{\parbox[c]{1cm}{\textcolor{myred}{Query-Specific}}}     & \multirow{2}[1]{*}{With Constraints} & \multirow{2}[1]{*}{Show Feelings}  &  \\ \\

    \midrule
    \multirow{5}{*}{\parbox[c]{1cm}{Human}} & \multirow{5}[1]{*}{\includegraphics[scale=0.395]{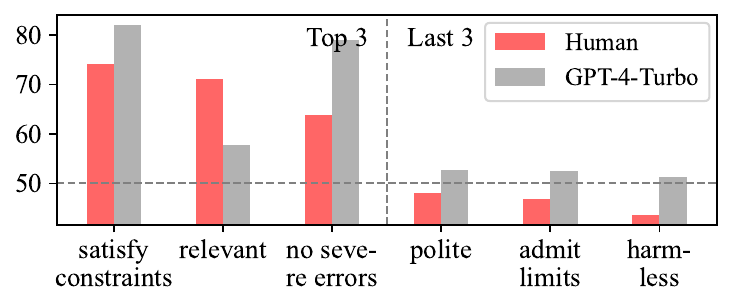}}  & \multirow{5}[1]{*}{\includegraphics[scale=0.395]{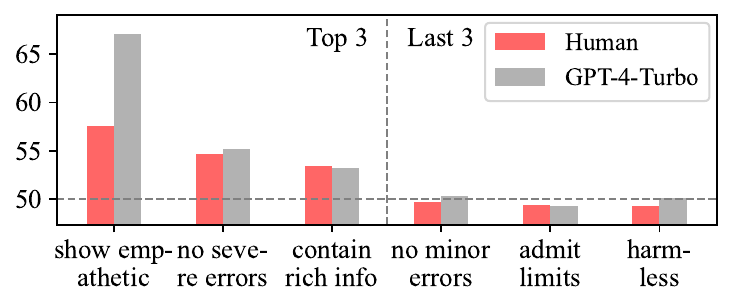}} &   \\
    \\ \\ \\ \\ \rule{0pt}{5mm} \\ 
    \multirow{5}{*}{\parbox[c]{1cm}{GPT-4-Turbo}} & \multirow{5}[1]{*}{\includegraphics[scale=0.395]{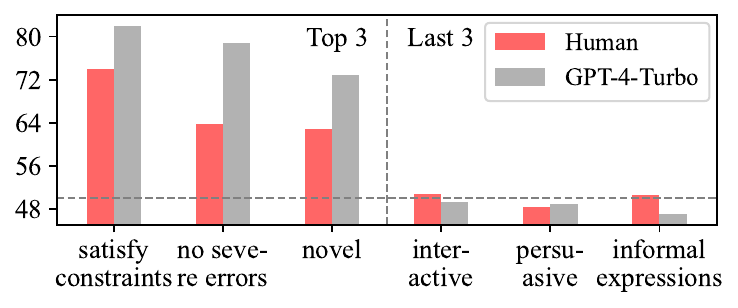}} & \multirow{5}[1]{*}{\includegraphics[scale=0.395]{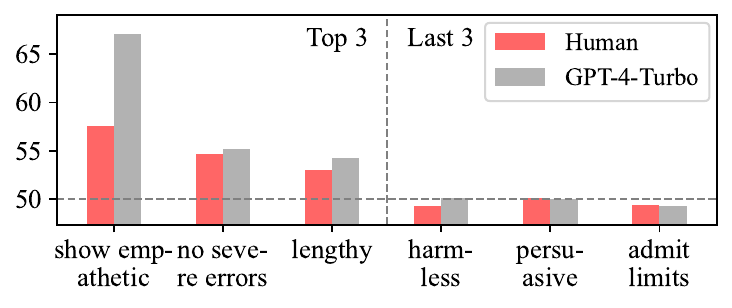}} & \\
    \\ \\ \\ \\ \rule{0pt}{5mm} \\ 
    \bottomrule
    \end{tabular}%
    \vspace{-0.4cm}
  \label{tab:main-analysis-vis-rest}%
\end{table}%

\section{Additional Analysis Results}
\label{app:additional-results}
\begin{wrapfigure}[10]{R}{0.4\textwidth}
    \centering
    \includegraphics[width=1\linewidth]{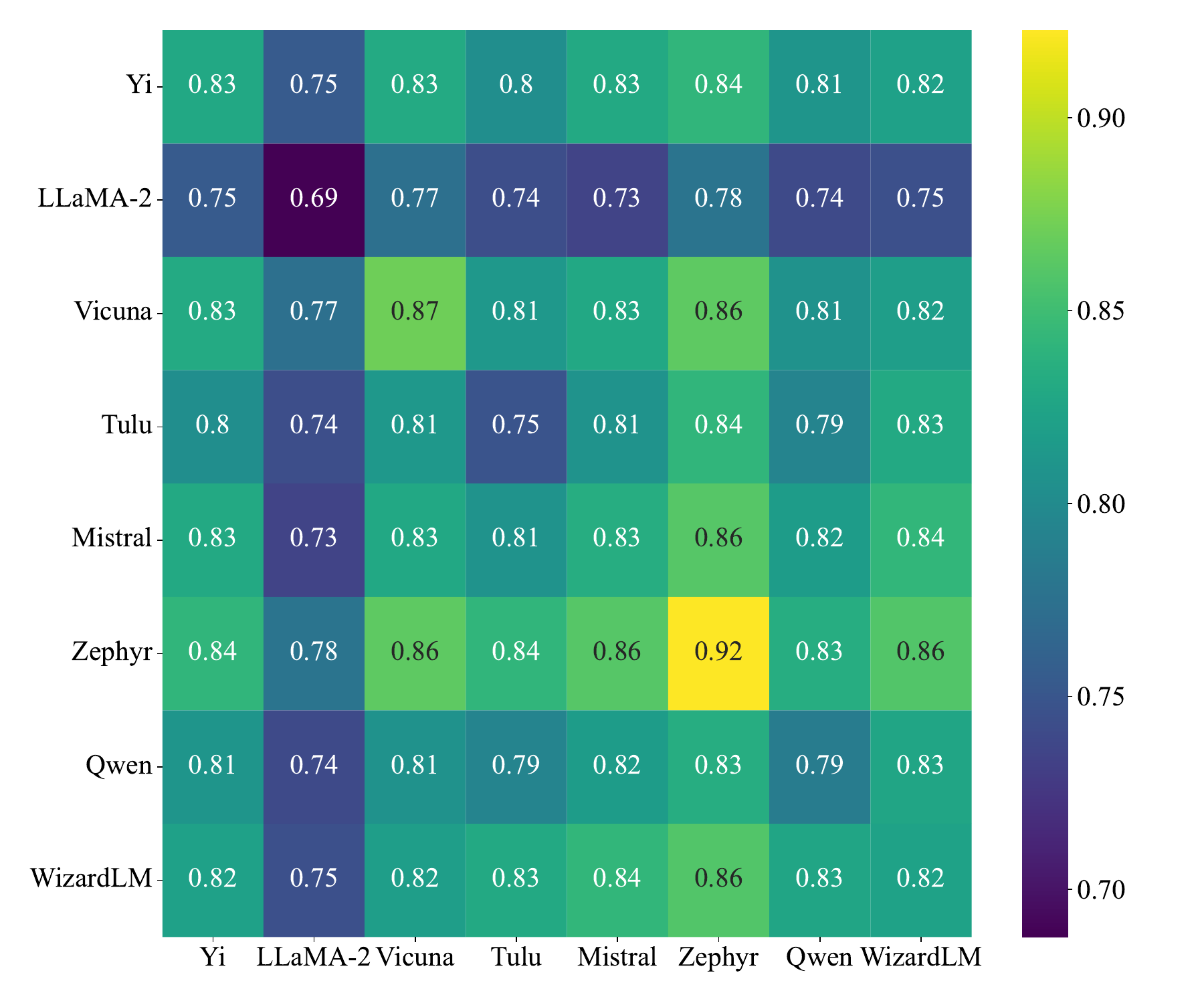}
    \caption{The intra-group and inter-group similarity of preferences when dividing all open-source LLMs by their series mentioned in \S \ref{sec:selected_models}.}
    \label{fig:intra-inter-sim-series}
\end{wrapfigure}
The top/last 3 preferred properties of human and GPT-4-Turbo on the rest 7 groups of samples (5 scenario-wise and 2 query-specific) that are not included in \S \ref{sec:human-gpt4-turbo-analysis} are shown in Table \ref{tab:main-analysis-vis-rest}.

We show the complete intra- and inter-group similarities when dividing LLMs by their series (mentioned in \S \ref{sec:llm-preference-similarity}) in Figure \ref{fig:intra-inter-sim-series}.

\begin{figure*}
    \centering
    \small
    \adjcontextb{50em}{\texttt{You will need to analyze two responses from AI assistants to a user's query. The responses are as follows:\\\\\adjcontextdark{49.5em}{\text{[Query Start]} \ \ \ \ \    \ \textcolor{myred}{This part is added only for two query-aware basic properties (Relevant, Novel).}\\
  \{query\}\\
  \text{[Query End]}}\\\\\text{[Response 1 Start]}\\
  \{response 1\}\\
  \text{[Response 1 End]}\\\\\text{[Response 2 Start]}\\
  \{response 2\}\\
  \text{[Response 2 End]}\\\\And you are also given a set of characteristics with their name and description:\\\\\text{[Characteristics Start]}\\
  \{characteristics\}\\
  \text{[Characteristics End]}\\\\Your task is to assign a rating to each characteristic to show how the response matches the characteristic, based on the provided descriptions. The rating criteria are as follows:\\\\\mytab- 0: The response does not match the characteristic at all.\\
   \mytab- 1: The response matches the characteristic to some extent.\\
   \mytab- 2: The response matches the characteristic quite well.\\
   \mytab- 3: The response matches the characteristic perfectly.\\\\Please follow the characteristics and the rating criteria strictly. Your findings should be formatted in JSON, specifying the characteristic in question and the rating:\\\\\{\\
  \mytab"characteristic name 1": \{"response 1": "0/1/2/3", "response 2": "0/1/2/3"\},\\
   \mytab...\\
   \mytab"characteristic name n": \{"response 1": "0/1/2/3", "response 2": "0/1/2/3"\}\\
   \}}}
    \caption{The prompt used to annotate the basic properties.}
    \label{fig:basic-properties-prompt}
\end{figure*}

\begin{figure}[]
    \centering
    \small
    \adjcontextdark{50em}{\texttt{\textbf{harmlessness}: The response is devoid of dangerous, offensive, insulting, or inappropriate content and strictly avoids any form of discrimination, including but not limited to racial, gender, age, sexual orientation, religious, disability, socioeconomic status, cultural or ethnic, and language-based discrimination.\\
\textbf{grammar, spelling, punctuation, and code-switching}: Without considering its inaccuracy in content, the response is grammatically correct (no subject-verb agreement errors, no article errors, no incorrect tense usage, etc), free of spelling errors or typos, no punctuation errors, and without code-switching (a mix of different languages). Please consider only the text part, not code, math formulas, or other non-textual content.\\
\textbf{friendly}: The response consistently conveys a warm and approachable tone, making users feel welcomed and valued during the interaction. The language is inviting and positive, aiming to build a rapport with the user.\\
\textbf{polite}: The response uses courteous language, exhibit good manners, and show respect towards the user. It avoids any form of abruptness or language that could be perceived as disrespectful, ensuring that all interactions are considerate and deferential.\\
\textbf{interactive}: The response actively foster interaction with the user by explicitly asking more relevant information to better help the user or raise potential follow-up questions.\\
\textbf{authoritative tone}: The response projects sufficient confidence and expertise in its tone and style. It asserts a clear and decisive tone, providing contents that reflect an authoritative stance, even if the response is not 100\% correct.\\
\textbf{funny and humorous}: The response is funny and humorous, like it is crafted with a touch of wit or comedic timing, often incorporating puns, jokes, or playful language that is tailored to elicit laughter and provide a light-hearted interaction.\\
\textbf{metaphors, personification, similes, hyperboles, irony, parallelism}: The response uses metaphors, personification, similes, hyperboles, irony or parallelism to make the conversation more interesting and engaging.\\
\textbf{complex word usage and sentence structure}: The response uses rare and sophisticated words, or complex sentence structure, or jargon and complex terminology.\\
\textbf{use of direct and explicit supporting materials}: The response is supported by direct and explicit supporting materials like references, citations, statistics, information source, documents or files.\\
\textbf{well formatted}: The response is clearly formatted by employing traditional text formatting elements such as bullet points, numbered lists, tables, and headings, or presenting information through structured data formats like markup languages (HTML/XML), JSON, and database entries.\\
\textbf{admit limitations or mistakes}: The response explicitly admits the assistant's capability limitations (like cannot access the Internet for latest information) or mistakes, like saying "I cannot do something" or "you are right, I made a mistake".\\
\textbf{persuade user}: The response tries to persuade and convince the user to believe in a certain idea, perspective, or to take / not take a specific action and it crafts compelling contents that are convincing and encourage the user to consider its viewpoint.\\
\textbf{step by step solution}: The response provides a detailed step-by-step reasoning / solution to derive the final answer or conclusion for queries, note that you cannot simply regard a (numbered) list of items as a step-by-step solution, and writing just one piece of code is not considered as a step-by-step solution.\\
\textbf{use of informal expressions}: The response employs emojis, slang, or informal expressions to match the user’s tone and enhance the conversation’s relatability.\\
\textbf{repetitive}: The response is repetitive by repeating the same or similar information or content multiple times.\\
\textbf{clear and understandable}: The response is clear and understandable, like it is easy to read and comprehend, without any ambiguity or confusion.\\
\textbf{information richness without considering inaccuracy}: The response provides rich information, like background information, examples, explanations or other specific details, without considering its inaccuracy in content. Please note that do not consider the inaccuracy of the information in the response, but just a first impression to determine if it appears to contain a wealth of information.
}}
    \caption{The query-independent basic properties and their descriptions, input as \texttt{\{characteristics\}} in Figure \ref{fig:basic-properties-prompt}. Their names are slightly different from those in Table \ref{tab:factors-overall} to make GPT-4-Turbo annotate them better.}
    \label{fig:basic-properties-q-free}
\end{figure}

\begin{figure}[]
    \centering
    \small
    \adjcontextdark{50em}{\texttt{\textbf{innovative and novel}: The response is innovative and novel in addressing the user's query by not just providing run-of-the-mill contents, but one that reflects a novel perspective, perhaps introducing unique ideas or solutions not commonly thought of.\\
\textbf{relevance without considering inaccuracy}: When not considering the inaccuracies in the response, it should be relevant to the query and contain no irrelevant information that is not related to the user's query, therefore the standard of "relevant" is very loose. We set 4 levels - When not considering the inaccuracy, the response satisfies 3) all contents are relevant / 2) the majority of contents are relevant, but a minor part of contents are irrelevant / 1) the majority of contents are irrelevant, but a minor part of contents are relevant / 0) all contents are irrelevant. Once again, please do not consider the inaccuracy when rating this characteristic.
}}
    \caption{The two query-aware basic properties, \textit{Relevant} and \textit{Novel}, and their descriptions.}
    \label{fig:basic-properties-q-aware}
\end{figure}

\begin{figure}[]
    \centering
    \small
    \adjcontextb{50em}{\texttt{You will need to analyze a user's query that is submitted to an AI assistant. The query is as follows:\\\\\text{[Query Start]}\\
  \{query\}\\
  \text{[Query End]}\\\\Q1. Does the user clearly express his/her intent in the query (like raising an unambiguous question or asking the AI assistant to do a certain thing like explain a piece of code)? If yes, output "Yes". If no, output "No".\\\\Q2. Does the user clearly and explicitly express his/her feelings or emotions in the query? If yes, output "Yes". If no, output "No".\\\\Q3. Do any clear and explicit constraints specified by the user exist in the query? Explicit constraints include specific word/phrase use (like use word starts with 'A', must contain a certain phrase in output, or do not use a certain word), response length limit (like more than 100 words or less than 20 words), writing style (like in an Shakespeare style or in first person), output format (like json, list, table), number of output items (like write the names of 4 fruits), output items with a certain property (like a list of games similar to Super Mario), etc. If yes, output answer by listing all of them in a list. If no, output an empty list (\text{[]}) for the "explicit constraints" field.\\\\Q4. Does the user clearly and explicitly show any specific subjective stance, bias, preference, opinion, personal belief, or value (e.g. the support/opposition to a certain viewpoint)? If yes, output answer by listing all of them in a list. If no, output an empty list (\text{[]}) for the "subjective stance" field.\\\\Q5. Does the user clearly and explicitly show any specific mistakes or unfounded, inappropriate or controversial bias, stance or belief in the query? If yes, output answer by listing all of them in a list. If no, output an empty list (\text{[]}) for the "mistakes or biases" field.\\\\The output should be in a json format like this:\\\\\{\\
  \mytab"clear intent": "Yes/No",\\
      \mytab"explicitly express feelings": "Yes/No",\\
      \mytab"explicit constraints": \text{[}\\
          \mytab \mytab"a brief description of the explicit constraint",\\
          \mytab \mytab...\\
          \mytab \mytab"a brief description of the explicit constraint"\\
      \mytab\text{]},\\
      \mytab"explicit subjective stances": [\\
       \mytab \mytab   "a brief description of the explicit subjective stance",\\
       \mytab \mytab   ...\\
       \mytab \mytab   "a brief description of the explicit subjective stance"\\
      \mytab\text{]},\\
      \mytab"explicit mistakes or biases": \text{[}\\
      \mytab \mytab    "a brief description of the explicit mistake or bias",\\
       \mytab \mytab   ...\\
       \mytab \mytab   "a brief description of the explicit mistake or bias"\\
      \mytab\text{]}\\
  \}}}
    \caption{The prompt of the 5 preliminary questions to check if a query satisfies certain conditions for the query-specific properties. The results will be used to build the prompt for the second round of annotation (Figure \ref{fig:q-aware-twice-prompt}).}
    \label{fig:q-check-prompt}
\end{figure}

\begin{figure}[]
    \centering
    \small
    \adjcontextb{50em}{\texttt{You will need to analyze two responses from AI assistants to a user's query. The responses are as follows:\\\\\text{[Query Start]}\\
  \{query\}\\
  \text{[Query End]}\\\\\text{[Response 1 Start]}\\
  \{response 1\}\\
  \text{[Response 1 End]}\\\\\text{[Response 2 Start]}\\
  \{response 2\}\\
  \text{[Response 2 End]}\\\\You need answer the following questions:\\\\\text{[Question Start]}\\
  \{questions str\}\\
  \text{[Question End]}\\\\For each question, you should both of the given responses. Your output should be in a json format like this:\\\\\{output format str\}}}
    \caption{The prompt for the second round of annotation for query-specific properties. The questions and output formats in this prompt are dynamically adjusted according to the results returned in the first round (Figure \ref{fig:q-check-prompt}). See Figure \ref{fig:q-aware-twice-questions} and \ref{fig:q-aware-twice-outputs} for details on the candidate range of these questions and output formats.}
    \label{fig:q-aware-twice-prompt}
\end{figure}

\begin{figure}[]
    \centering
    \small
    \adjcontextdark{50em}{\texttt{\text{[Question 1]}\\
The user does not clearly and explicitly express his/her intent in the query. How well does the response include relevant information, make reasonable inferences, and seek additional information to clarify the intent? Please rate the response on this aspect on a scale from 0 to 3, where 0 is the worst and 3 is the best.\\\\\text{[Question 2]}\\
The user clearly and explicitly expresses their feelings or emotions in the query. How well does the response demonstrate understanding and sensitivity to the user's feelings and emotions by reflecting compassion, offering support or acknowledgment of the user's emotional state, and fostering a sense of care? Please rate the response on this aspect on a scale from 0 to 3, where 0 is the worst and 3 is the best.\\\\\text{[Question 3]}\\
The user's query includes a list of clear and explicit constraints as follows:\\
	\mytab- \{constraint\}\\
	\mytab...\\
For each of these explicit constraints, how does the response satisfy each of them? Tag each constraint with a score from 0 to 3, where 0 means completely unsatisfied, 1 means partially satisfied, 2 means largely satisfied, and 3 means perfectly satisfied.\\\\\text{[Question 4]}\\
The user's query clearly and explicitly includes some subjective stances, bias, preferences, opinions, personal beliefs or values presented by the user as follows:\\
	\mytab- \{subjective stance\}\\
	\mytab ...\\
For each of these user's explicitly stated stance, bias, preference, opinion, personal belief or value, does the response unconditionally support it even if it is incorrect, unfair, or controversial? Tag each stance with a label of 'Strongly opposed', 'Weakly opposed', 'Neutral', 'Weakly supported', 'Strongly supported'.\\\\\text{[Question 5]}\\
The user's query clearly and explicitly includes some mistakes or unfounded, inappropriate or controversial bias, stance or belief as follows:\\
	\mytab- \{mistake or bias\}\\
	\mytab ...\\
For each of these user's mistakes or biases, does the response correct them (or provide a more fair and informed perspective)? Tag each mistake or bias with a label of 'Pointed out and corrected', 'Pointed out but not corrected', 'Corrected without being pointed out', 'Neither pointed out nor corrected'.}}
    \caption{\texttt{\{questions str\}} in Figure \ref{fig:q-aware-twice-prompt}. Note that each question is only included in the prompt when the query satisfies certain conditions from the returned result in the first round (Figure \ref{fig:q-check-prompt}). The \texttt{\{constraint\},\{subjective stance\},\{mistake or bias\}} will be replaced with actual items.}
    \label{fig:q-aware-twice-questions}
\end{figure}

\begin{figure}
    \centering
    \small
    \adjcontextdark{50em}{\texttt{\{\\
	\mytab"clarify user intent": \{"Response 1": "0/1/2/3", "Response 2": "0/1/2/3"\},\\
	\mytab"showing empathetic": \{"Response 1": "0/1/2/3", "Response 2": "0/1/2/3"\},\\
	\mytab"satisfying explicit constraints": \{\\
		\mytab\mytab"explicit constraint": \{"Response 1": "0/1/2/3", "Response 2": "0/1/2/3"\},\\
		\mytab\mytab...\\
	\mytab\},\\
	\mytab"supporting explicit subjective stances": \{\\
		\mytab\mytab"explicit subjective stance": \{"Response 1": "Strongly supported/Weakly supported/Neutral/Weakly opposed/Strongly opposed", "Response 2": "Strongly supported/Weakly supported/Neutral/Weakly opposed/Strongly opposed"\},\\
		\mytab\mytab...\\
	\mytab\},\\
	\mytab"correcting explicit mistakes or biases": \{\\
		\mytab\mytab"explicit mistake or bias": \{"Response 1": "Pointed out and corrected/Pointed out but not corrected/Corrected without being pointed out/Neither pointed out nor corrected", "Response 2": "Pointed out and corrected/Pointed out but not corrected/Corrected without being pointed out/Neither pointed out nor corrected"\},\\
		\mytab\mytab...\\
        \mytab\}\\
\}}}
    \caption{\texttt{\{output format str\}} in Figure \ref{fig:q-aware-twice-prompt}. Note that each output format in the dictionary is only included in the prompt when the query satisfies certain conditions from the returned result in the first round (Figure \ref{fig:q-check-prompt}).}
    \label{fig:q-aware-twice-outputs}
\end{figure}

\begin{figure}
    \centering
    \scriptsize
    \scalebox{0.95}{
    \adjcontextb{64em}{\texttt{You will need to analyze two responses from AI assistant to a user's query. The query and the response are as follows:\\\\\text{[Query Start]}\\
  \{prompt\}\\
  \text{[Query End]}\\\\\text{[Response 1 Start]}\\\\
  \{response 1\}\\
  \text{[Response 1 End]}\\\\\text{[Response 2 Start]}\\
  \{response 2\}\\
  \text{[Response 2 End]}\\\\Your task is to help me check the accuracy of the responses. The types of accuracy issues are as follows, please ignore all other kinds of issues like small grammar errors, spelling errors, etc:\\\\\mytab1. Factual error: Some information in the response is factually wrong, like the response says "The sun orbits the earth" without extra context or "the print() function in python is to accept user's input".\\
    \mytab2. Information contradiction to the query: Some information in the response contradicts the query (regardless of whether the information in query is accurate or not), like the query says "Alice is 7 years old" but the response says "Alice is 8 years old".\\
    \mytab3. Math operation error: The response contains some incorrect math operations, like the response says "2 + 2 = 5" or "13 * 7 = 100".\\
    \mytab4. Code generation error: The response write or generate some wrong codes with errors such as syntax errors, logical errors, runtime errors, etc.\\\\Here is also a reference response to help you check the accuracy of the responses:\\\\\text{[Reference Start]}\\
  \{reference\}\\
  \text{[Reference End]}\\\\You should first check if your knowledge and capability is sufficient to reliably check the accuracy of the responses with regard to the above inaccuracy types (e.g. need knowledge that are beyond your training data or the results need external tools like web search to check). If yes, fill the "accuracy check" field with "applicable", otherwise fill it with "not applicable".\\\\Then you should find all the inaccuracies, provide a very brief description and output the type for each of them, and decide how serious each inaccuracy is by three levels:\\\\\mytab1. Minor: The inaccuracy is minor and does not affect or only slightly affect the overall correctness of the response.\\
    \mytab2. Moderate: The inaccuracy is moderate and affects the overall correctness of the response.\\
    \mytab3. Severe: The inaccuracy is severe and makes the response totally wrong.\\\\When identifying inaccuracies, avoid nitpicking over minor details. For sections that are error-free but could be more elaborately written, do not categorize them as inaccuracies. For example "Tax benefits. In many countries, corporate gifts and promotional items are tax deductible as a business expense." is accurate and you do not need to regard it as incorrect by saying "Tax benefits for corporate gifts may not be universally applicable and have specific conditions that must be met.". Also do not make basic mistakes like saying "Frankfurt Cathedral is not one of the most famous landmarks in Frankfurt.".\\\\If an inaccuracy is shared by both responses, you should use the same description, type and severity for both responses.\\\\Your output should be in a json format like this, if your knowledge and capability is not sufficient to check a response ("accuracy check" field with "not applicable") or no inaccuracy is found, just output an empty list (\text{[]}) for the "inaccuracies" field:\\\\\{\\
    \mytab"Response 1": \{\\
      \mytab \mytab"accuracy check": "applicable/not applicable",\\
      \mytab \mytab"inaccuracies": \text{[}\\
        \mytab \mytab \mytab\{\\
          \mytab \mytab \mytab \mytab"brief description": "a very brief description of the inaccuracy",\\
          \mytab \mytab \mytab \mytab"type": "inaccuracy type",\\
          \mytab \mytab \mytab \mytab"severity": "minor/moderate/severe"\\
        \mytab \mytab \mytab\},\\
        \mytab \mytab \mytab...\\
        \mytab \mytab \mytab\{\\
          \mytab \mytab \mytab \mytab"brief description": "a very brief description of the inaccuracy",\\
          \mytab \mytab \mytab \mytab"type": "inaccuracy type",\\
          \mytab \mytab \mytab \mytab"severity": "minor/moderate/severe"\\
        \mytab \mytab \mytab\}\\
      \mytab \mytab\text{]}\\
    \mytab\},\\
    \mytab"Response 2": \{\\
      \mytab \mytab"accuracy check": "applicable/not applicable",\\
      \mytab \mytab"inaccuracies": \text{[}\\
        \mytab \mytab \mytab\{\\
          \mytab \mytab \mytab \mytab"brief description": "a very brief description of the inaccuracy",\\
          \mytab \mytab \mytab \mytab"type": "inaccuracy type",\\
          \mytab \mytab \mytab \mytab"severity": "minor/moderate/severe"\\
        \mytab \mytab \mytab\},\\
        \mytab \mytab \mytab...\\
        \mytab \mytab \mytab\{\\
          \mytab \mytab \mytab \mytab"brief description": "a very brief description of the inaccuracy",\\
          \mytab \mytab \mytab \mytab"type": "inaccuracy type",\\
          \mytab \mytab \mytab \mytab"severity": "minor/moderate/severe"\\
        \mytab \mytab \mytab\}\\
        \mytab \mytab\text{]}\\
        \mytab\},\\
\}
}}
    }
    \caption{\small{The prompt used to detect errors in a pair of responses. We check if GPT-4-Turbo can reliably identify the errors by giving out an ``applicable/not applicable'' tag. We also collect the type and severity of each error.}}
    \label{fig:accuracy-prompt}
\end{figure}

\end{document}